\documentclass[preprint,12pt]{elsarticle}

\usepackage{amssymb,graphicx,caption,amsmath,mathrsfs}
\usepackage{bbm,booktabs,float,multirow,multicol,pifont,mathtools}
\usepackage{soul,color,xcolor}
\biboptions{sort&compress}
\usepackage{algorithm}
\usepackage{algorithmicx}
\usepackage{algpseudocode}
\usepackage{bm}
\usepackage{hyperref}
\hypersetup{hidelinks,
	colorlinks=true,
	allcolors=black,
	pdfstartview=Fit,
	breaklinks=true}

\journal{ELSEVIER}
\soulregister\cite7
\soulregister\ref7
\begin{document}

\begin{frontmatter}



\title{Tackling Data Heterogeneity in Federated Learning through Knowledge Distillation with Inequitable Aggregation}


\author[a]{Xing Ma\corref{cor1}}
\ead{maxxldnc@163.com}

\cortext[cor1]{Corresponding author}
\affiliation[a]{organization={School of Computer Science and Technology},
            addressline={Harbin University of Science and Technology}, 
            city={Harbin},
            postcode={150080}, 
            country={China}}

\begin{abstract}
Federated learning aims to train a global model in a distributed environment that is close to the performance of centralized training. However, issues such as client label skew, data quantity skew, and other heterogeneity problems severely degrade the model's performance. Most existing methods overlook the scenario where only a small portion of clients participate in training within a large-scale client setting, whereas our experiments show that this scenario presents a more challenging federated learning task.  Therefore, we propose a Knowledge Distillation with teacher-student Inequitable Aggregation (KDIA) strategy tailored to address the federated learning setting mentioned above, which can effectively leverage knowledge from all clients. In KDIA, the student model is the average aggregation of the participating clients, while the teacher model is formed by a weighted aggregation of all clients based on three frequencies: participation intervals, participation counts, and data volume proportions. During local training, self-knowledge distillation is performed. Additionally, we utilize a generator trained on the server to generate approximately independent and identically distributed (IID) data features locally for auxiliary training. We conduct extensive experiments on the CIFAR-10/100/CINIC-10 datasets and various heterogeneous settings to evaluate KDIA. The results show that KDIA can achieve better accuracy with fewer rounds of training, and the improvement is more significant under severe heterogeneity. Code: \href{https://github.com/maxiaum/KDIA}{https://github.com/maxiaum/KDIA}.
\end{abstract}




\end{frontmatter}


\section{Introduction}\label{sec1}
In recent years, deep learning has been widely applied in industrial production and people's daily lives, and the nourishment of deep learning—big data has been fully utilized. However, with the growing emphasis on data privacy, the problem of data silos has severely hindered the further development of deep learning \cite{zhang2021survey}. To address this, McMahan et al. proposed the Federated Learning framework and the FedAvg algorithm \cite{mcmahan2017communication}, where clients can collaboratively train a global model with the server without sharing their local data. FedAvg can achieve near-centralized training performance with independent and identically distributed (IID) data, but in heterogeneous environments, the method faces negative impacts on model convergence and accuracy.\\
Data heterogeneity is a widely studied problem in federated learning, which mainly includes issues like quantity skew, distribution-based label skew, and quantity-based label skew \cite{zhao2018federated, zhu2021federated, li2022federated}. The first two are unique heterogeneous scenarios in the federated learning framework. In experiments, these heterogeneous scenarios are often simulated by applying Dirichlet distributions or artificially partitioned data. Data heterogeneity makes local models more "personalized", but this kind of "personalization" is not beneficial to the global model. Instead, it causes local models to gradually deviate from the update direction of the target model (the centralized model) during the training process, resulting in a decrease in the accuracy and convergence speed of the global model. Existing methods attempt to solve the problem of data heterogeneity from two aspects: local update and global aggregation. MOON \cite{li2021model} adopts the idea of contrastive learning to narrow the difference between local models and the global model. FedGKD \cite{yao2021local} uses knowledge distillation to guide the training of local models. FedDisco \cite{ye2023feddisco} adjusts the performance of the global model by dynamically calculating the aggregation weights. However, the above methods have two deficiencies: (1) They focus on the scenario where all small-scale clients participate in the training, while ignoring the natural characteristic of federated learning, which is "random client sampling"; (2) There is no effective utilization of clients that do not participate in the training in each training session. Aiming at the above problems, this paper conducts research on the federated scenario with a low client participation rate under a large scale of clients. Through experiments, it is found that this scenario intensifies the impact of data heterogeneity on the model to a certain extent. Increasing the client sampling ratio can indeed improve the model performance, but on the other hand, it increases the possibility of communication overhead and task progress delay \cite{ribero2007communication, fu2023client}. Therefore, the above scenario is a more challenging task that is more in line with the characteristics of federated learning. \\
In this paper, we propose a communication-efficient method that utilizes all clients for training—Knowledge Distillation with teacher-student Inequitable Aggregation (KDIA). Unlike the traditional approach of obtaining the teacher model through a historical global model or simple weighted aggregation of all client data, we introduce \textit{triFreqs} (triple-frequencies) to weight the aggregation across all clients, considering three frequencies: client participation intervals, participation counts, and data volume proportions. This approach helps avoid the issues of slow teacher model updates leading to ineffective guidance and fast updates causing knowledge forgetting. Additionally, we adopt the generation method of FedGen \cite{zhu2021data} to locally generate data features with approximately uniform class distribution to assist in training. Finally, we provide a convergence analysis of the proposed method.
Our main contributions are as follows:
\begin{itemize}
    \item[$\bullet$] We experimentally explore the impact of client numbers and sampling ratios on federated learning, and based on the experimental results, we conclude that the data heterogeneity scenario under large-scale clients and low sampling ratios is more challenging.
    \item[$\bullet$] We propose a knowledge distillation with teacher-student inequitable aggregation method, where the server uses a geometric mean approach to average client participation intervals, participation counts, and data volume proportions to obtain weights for different clients, and then aggregates them to form the teacher model. This model then guides the local model training process, where only a subset of students participate.
    \item[$\bullet$] We train a class-balanced conditional generator on the server, and then generate approximate IID data locally for auxiliary training, while also performing data quantity augmentation for clients with insufficient data.
    \item[$\bullet$] Through extensive experiments, such as test accuracy, communication efficiency, and ablation studies, we validate the effectiveness of KDIA.
\end{itemize}
\section{Related Work}
\subsection{Federated Learning}
Federated learning is a popular research direction. Due to its distributed framework design, data heterogeneity and device heterogeneity are inherent limitations. This paper mainly focuses on the issue of data heterogeneity. Currently, the methods to address data heterogeneity in federated learning can be divided into two main directions: client-based local update methods and global-based model aggregation methods.

\textbf{Client-based local update methods.} These methods primarily involve adding regularization terms or penalty terms to the local training objective function to limit local updates and mitigate the differences between clients. FedProx \cite{li2020federated} introduces an $L_2$ regularization term in the objective function to reduce the difference between local updates and the global model during training. MOON \cite{li2021model} adopts a contrastive learning approach, shrinking the gap between the current local model and the global model during local training, while increasing the gap between the current local model and the previous local model. FedRS \cite{li2021fedrs} identifies that label skew (especially missing classes) primarily affects the last layer of the classifier, and proposes Restricted Softmax to limit the weight updates for missing classes. FedUFO \cite{zhang2021federated} proposes a unified feature representation learning and optimization goal alignment to reduce the differences between the global and local models. FedDC \cite{gao2022feddc} introduces a local training model offset decoupling and correction method, similar to local parameter momentum adjustment, to gradually alleviate the deviation between local updates and historical parameters. FedDecorr \cite{shi2022towards} proposes a dimension collapse theory under federated heterogeneity, measuring the degree of collapse through the singular value distribution of the covariance matrix of model parameters, and adds a singular value variance regularization term in the objective function to mitigate dimension collapse. FedUV \cite{son2024feduv} observes constant singular values in classifiers under data heterogeneity and simulates an IID setup by promoting the variance of the predicted dimensional probability distribution. FedCOG \cite{yefake} employs both global and local models to train data generators locally for auxiliary data synthesis in knowledge distillation, its heavy dependence on the model performance leads to suboptimal generation quality. Although these methods have achieved good results in certain federated environments, they overlook the potential knowledge source from other clients when only a subset of clients participate in the training.

\textbf{Global-based model aggregation methods.} These methods primarily involve adopting certain strategies on the server side to mitigate the impact of differences between client models. FedAvgM \cite{hsu1909measuring} uses a momentum-based sampling approach to adjust or reduce the update magnitude of the global model, making it more stable. FedNova \cite{wang2020tackling} focuses on the issue of data distribution shift across clients. Due to differences in local update times, the aggregation rounds and convergence rates for different clients may vary, leading to model drift. To address this, FedNova dynamically adjusts the aggregation weights based on the number of local updates. FedFTG \cite{zhang2022fine} uses a data generation method similar to FedGen \cite{zhu2021data}, generating auxiliary data on the server side, and then training the aggregated global model with this data to reduce the gap between the global and local models. FedDisco \cite{ye2023feddisco} introduces an adaptive discrepancy aggregation method, which dynamically calculates the aggregation weights for different clients based on local data volume and the discrepancy between clients and the server. These methods, however, only consider the aggregation of the models from selected clients and do not fully leverage the knowledge from other clients.
\subsection{Federated Knowledge Distillation}\label{subsec1}
Knowledge distillation is commonly used for model compression and knowledge transfer, making it well-suited for integration with federated learning \cite{hinton2015distilling, qin2024knowledge}. For instance, during local training, the teacher model can guide the student model’s updates. Knowledge distillation can be categorized in several ways: (1) offline distillation, online distillation, and self-distillation. In offline and online distillation, the teacher and student models have different structures. Offline distillation requires additional global data to train the teacher model in advance, while online distillation involves training and updating the teacher model together with the student model. Therefore, self-distillation is widely applied in federated learning. In self-distillation, the model serves as both the teacher and the student, allowing flexible selection between the two during the final model update. (2) feature-based distillation, relation-based distillation, and response-based distillation. This classification is primarily based on the source of the knowledge. Feature-based distillation focuses on the relationship between the feature maps obtained by the teacher and student models for the same input, and involves continuously matching these features during training. Relation-based distillation focuses on the relationships between the input layer, hidden layers, and output layer, considering both model weights and input data. Response-based distillation focuses solely on the output layer, where the student model matches or mimics the teacher’s output to achieve knowledge transfer. This method is simple and effective, which is why it is widely used in federated learning \cite{gou2021knowledge, DBLP:journals/corr/abs-2011-02367}.\\
In existing methods, FedGKD \cite{yao2021local} employs self-knowledge distillation, where the teacher model assists local training. The teacher model is obtained by averaging and aggregating multiple historical global models. However, averaging gives too much weight to the oldest global updates, which may conflict with the most recent updates. FedDM \cite{xiong2023feddm} uses data distillation to locally train synthetic data that can match the real data landscape. After uploading the synthetic data to the server, the server uses the aggregated data to train the global model. FedFed \cite{yang2024fedfed} utilizes feature distillation to classify local data features into performance-robust (features with little contribution to performance improvement) but information-sensitive features, and performance-sensitive (features with significant contribution to performance improvement) features. Performance-sensitive features, after being shared with differential privacy among clients, are used for local updates. However, none of these methods take all clients into account when aggregating the teacher model. FedKF \cite{zhou2022handling} uses self-knowledge distillation, where the teacher model is obtained by performing a quantity-weighted average across all clients. Although this method takes all clients into account during teacher model aggregation, aggregating based solely on data volume weights gives too much weight to clients that did not participate in training, leading to excessive focus on old knowledge.
\section{Notation and Preliminaries}
\textbf{Federated learning.} Assume the total number of clients participating in federated learning is $N$, forming the client set $S_a=\{c_1,c_2,\dots,c_N\}$. The data volume of each client is $N_k$ with the data set of client \textit{k} represented as $\bm{D}_k=\{(X_j,Y_j)\}_{j=1}^{N_k}$, where $X_j$ is the sample and $Y_j$ is the corresponding label. The client participation rate is $C\in(0,1.0]$, meaning the number of clients participating in each round of training is $K=N\times{C}$, and the set of participating clients in round $S_t=\{c_1,c_2,\cdots,c_K\}$. The global model is $\bm{\theta}_g=\langle\bm{\theta}_g^E,\bm{\theta}_g^C\rangle$, where $\bm{\theta}_g^E$ and $\bm{\theta}_g^C$ represent the feature extractor and the classifier, respectively. And in this paper, the convolutional part of the model is regarded as the extractor, while the linear part is treated as the classifier. The objective function of federated learning is:
\begin{equation}\label{eq1}
\min_{\langle\bm{\theta}_g^E,\bm{\theta}_g^C\rangle}\mathcal{L}\left(\langle\bm{\theta}_g^E,\bm{\theta}_g^C\rangle\right)=\sum_{k=1}^Kp_k\cdot\mathcal{L}_{CE}\left(\langle\bm{\theta}_k^E,\bm{\theta}_k^C\rangle\right),
\end{equation}
\begin{equation}\label{eq2}
    \mathcal{L}_{CE}(\langle\bm{\theta}_k^E,\bm{\theta}_k^C\rangle)=\mathbb{E}_{(X,Y)\sim \bm{D}_k}\ell_{CE}\left(\sigma\left(\left(\langle\bm{\theta}_k^E,\bm{\theta}_k^C\rangle;X\right)\right),Y\right)
\end{equation}
where $p_k=\frac{N_k}{\sum_{k\in S_t}N_k}$ is the weight of each client, and $\sum_{k\in S_t}p_k=1$. Eq. (\ref{eq2}) is the objective function of each client, where \textit{CE} is the cross-entropy loss function, and $\sigma$ is the softmax activation function. The federated learning model training process is as follows:
\begin{itemize}
    \item [(1)] The server initializes $\bm{\theta}_g$ and selects the client set $S_t$, then distributes $\bm{\theta}_g$ to the clients in $S_t$.
    \item [(2)] Each client in $S_t$ performs local updates using its own data and uploads $\langle\bm{\theta}_k^E,\bm{\theta}_k^C\rangle$ to the server.
    \item [(3)] The server aggregates the client models using Eq. (\ref{eq1}) to obtain the new $\bm{\theta}_g$.
    \item [(4)]These three steps are repeated until the model $\bm{\theta}_g$ converges. 
\end{itemize}
Our method follows the same process as described above, with the addition of teacher model aggregation and generator training at the server.

\textbf{Conditional generator.} The Conditional Generator (CG) is a commonly used data generation model \cite{zhu2021data, mirza2014conditional,wu2021fedcg}. Let the generator be denoted as $\bm{\theta}_{gen}$. The general process for CG to generate data is as follows:
\begin{itemize}
    \item[(1)] Sample noise data $\bm{\epsilon}{\sim}\mathcal{N}(0,I)$ from a Gaussian distribution, and sample labels $\bm{\tilde{Y}}{\sim}U(0,N_c-1)$ from a uniform distribution, where $N_c$ is the number of categories.
    \item[(2)] Input $\bm{\epsilon}$ and $\bm{\tilde{Y}}$ into $\bm{\theta}_{gen}$, and use $\bm{\theta}_{gen}$ to generate simulated features $\bm{\tilde{Z}}=\left(\bm{\theta}_{gen};(\bm{\epsilon},\bm{\tilde{Y}})\right)$ through multiple deconvolutions.
    \item[(3)] Input $\bm{\tilde{Z}}$ into the classifier $\bm{\theta}_*^C$ for training, during which $\bm{\theta}_*^C$ is frozen and only $\bm{\theta}_{gen}$ is updated.
\end{itemize}
The objective function is:
{\small
\begin{equation}\label{eq3}
\min_{\langle\bm{\theta}_{gen},\bm{\theta}_{*}^{C}\rangle}\mathbb{E}_{[\bm{\epsilon},\bm{\tilde{Y}}]}\left[\mathcal{L}_{CE}\left(\sigma\left(\frac{1}{K}\sum_{k=1}^{K}p_{k}\cdot\left(\langle\bm{\theta}_{gen},\bm{\theta}_{k}^{C}\rangle;(\bm{\epsilon},\bm{\tilde{Y}})\right)\right),\bm{\tilde{Y}}\right)\right].
\end{equation}
}
To enhance the diversity of the generated data and reduce the occurrence of identical data being generated from identical samples, a diversity regularization term is added to the objective function. Specifically, the noise data $\bm{\epsilon}$ and the simulated features $\bm{\tilde{Z}}$ are divided into two parts, and the following objective function is optimized:
\begin{equation}\label{eq4}
    \min_{\bm{\theta}_{gen}}\frac{|\bm{\epsilon}_1-\bm{\epsilon}_2|}{|\bm{\tilde{Z}}_1-\bm{\tilde{Z}}_2|+\varepsilon},
\end{equation}
where $\varepsilon>0$ to prevent the denominator in Eq. (\ref{eq4}) from being zero. The objective function aims to make similar samples generate different simulated features as much as possible.
\begin{table}
    \centering
        \caption{The table categorizes different methods based on three configurations. "\textit{large N/small C}" indicates experiments with a larger \textit{N} ($N\ge{100}$) and a smaller \textit{C} ($C\le{0.2}$); "\textit{small N/large C}" refers to experiments with a smaller \textit{N} ($N\le{30}$) and a larger \textit{C} ($C\ge{0.5}$); "\textit{mixed}" represents a mixed configuration.}
    \label{tab1}
    \resizebox{\linewidth}{!}{
    \begin{tabular}{c|c}
        \toprule
        Setting&Method\\
        \cline{1-2}
        \textit{large N/small C}&FedAvg \cite{mcmahan2017communication}, FedProx \cite{li2020federated},FedUFO \cite{zhang2021federated}, FedFTG \cite{zhang2022fine}\\
        \cline{1-2}
        \textit{mixed}&FedGen \cite{zhu2021data}, FedRS \cite{li2021fedrs}, FedDC \cite{gao2022feddc}, FedAvgM \cite{hsu1909measuring}\\
        &FedGKD \cite{yao2021local}, FedKF \cite{zhou2022handling}\\
        \cline{1-2}
        \textit{small N/large C}& FedNova \cite{wang2020tackling}, MOON \cite{li2021model}, FedDisco \cite{ye2023feddisco},  FedUV \cite{son2024feduv}, \\
        & FedDecorr \cite{shi2022towards}, FedCOG \cite{yefake}, FedDM \cite{xiong2023feddm}, FedFed \cite{yang2024fedfed}, \\
        \bottomrule
    \end{tabular}
    }
\end{table}
\section{Motivation}\label{s3}
Data heterogeneity is an inherent issue in federated learning. Existing methods for constructing data heterogeneity primarily involve Dirichlet partitioning (\textit{Dir($\beta$)}) and configurations where each client contains only a subset of classes. After the data is partitioned, experiments are typically conducted with a fixed number of clients \textit{N} and a fixed sampling ratio \textit{C}, which may somewhat overlook the impact of \textit{N} and \textit{C} on the method. Table \ref{tab1} summarizes the values of \textit{N} and C used in various methods. The "small \textit{N}/large \textit{C}" configuration sets a small \textit{N} and a sampling ratio $C=1.0$, which seems inconsistent with the characteristics of federated learning—particularly the sampling process of clients. When \textit{N} is large or there are transmission efficiency constraints, a smaller client sampling ratio is required to deal with stragglers \cite{fu2023client,li2021fedrs}. Therefore, we conducted experiments to explore the effects of the number of clients \textit{N} and the sampling ratio \textit{C} on the method and propose an effective approach that can utilize all clients for training.\\
\begin{figure}
    \begin{minipage}{0.23\linewidth}
		\vspace{3pt}
		\centerline{\includegraphics[width=\textwidth]{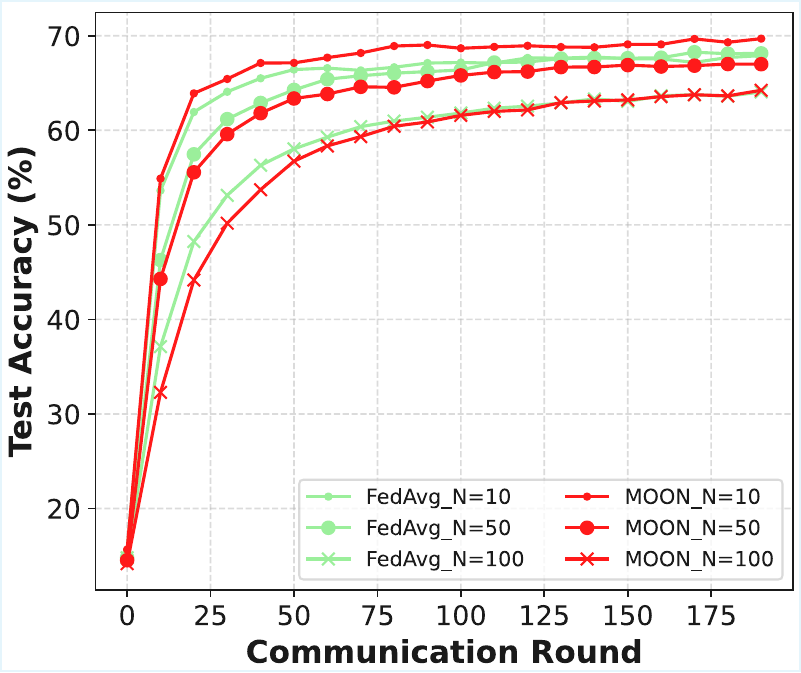}}
		\centerline{\footnotesize (a) Client number \textit{N}}
	\end{minipage}
	\begin{minipage}{0.23\linewidth}
		\vspace{3pt}
		\centerline{\includegraphics[width=\textwidth]{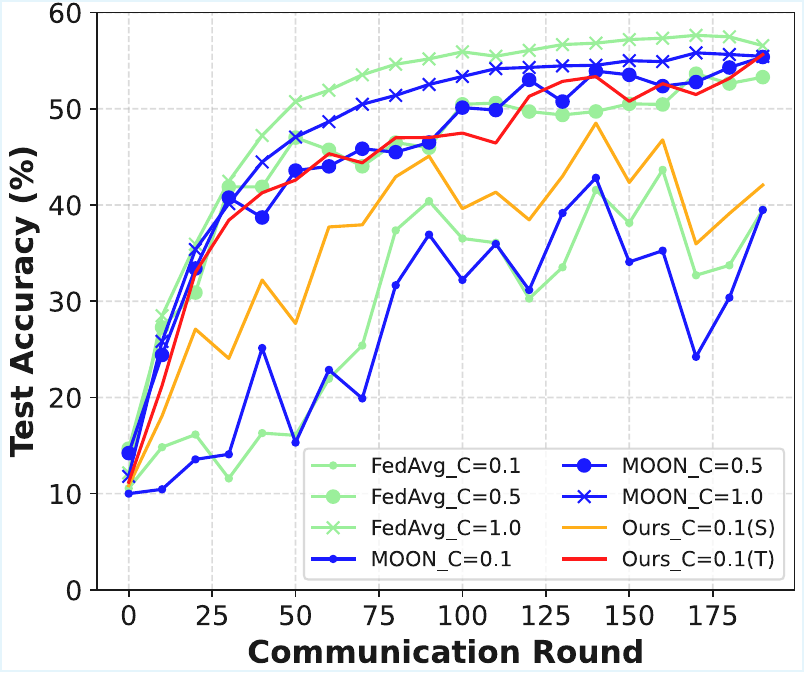}}
		\centerline{\footnotesize (b) $\beta=0.1$}
	\end{minipage}
	\begin{minipage}{0.23\linewidth}
		\vspace{3pt}
		\centerline{\includegraphics[width=\textwidth]{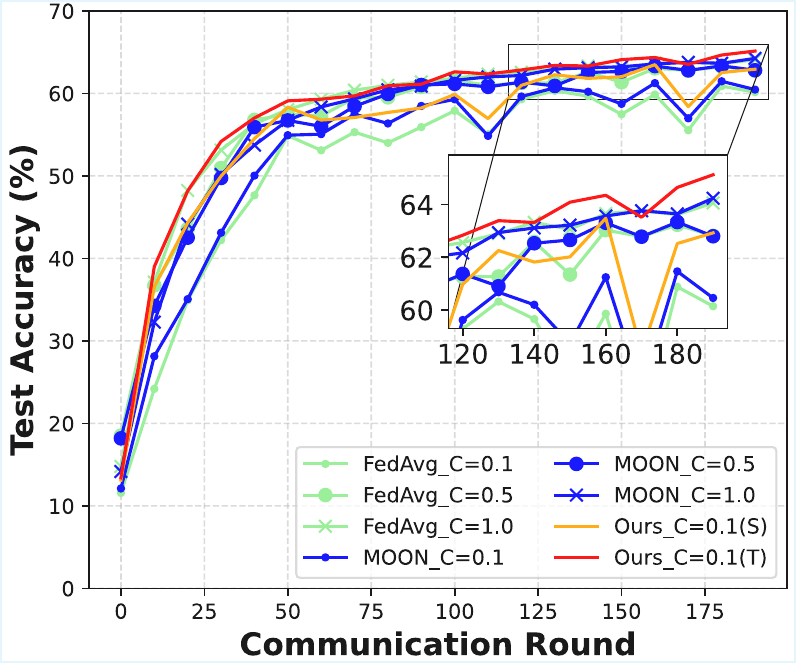}}
		\centerline{\footnotesize (c) $\beta=0.5$}
	\end{minipage}
        \begin{minipage}{0.23\linewidth}
		\vspace{3pt}
		\centerline{\includegraphics[width=\textwidth]{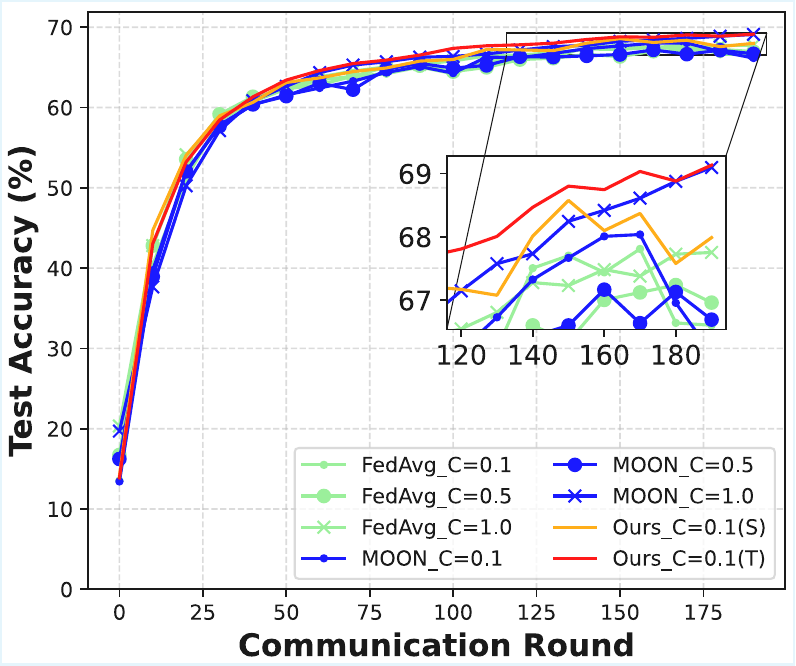}}
		\centerline{\footnotesize (d) $\beta=5.0$}
	\end{minipage}
	\caption{On the CIFAR-10 dataset, (a) shows the test accuracy curves of FedAvg and MOON with different numbers of clients \textit{N}; (b), (c), (d) show the test accuracy curves of FedAvg and MOON with a fixed \textit{N} under different levels of heterogeneity, while varying the client sampling ratio \textit{C}.}
	\label{fig1}
\end{figure}
In the Internet of Things, the number of clients is in the thousands. To facilitate the experiment, we define the number of clients that is greater than or equal to the number adopted by FedAvg as a large scale ($N\ge{100}$) \cite{mcmahan2017communication}. In exploring the impact of the number of clients \textit{N} and the sampling ratio \textit{C} on the method, we partitioned the data using \textit{Dir($\beta$)} and conducted experiments on FedAvg \cite{mcmahan2017communication} and MOON \cite{li2021model}. For \textit{N}, we fixed $C=1.0$, $\beta=0.5$ and varied $N=10/50/100$; For \textit{C}, we fixed $N=100$ (based on results from experiments with \textit{N}, and varied $C=0.1/0.5/1.0$. The experimental results are shown in Fig. \ref{fig1}.
According to Fig. \ref{fig1} (a), the test accuracy decreases as \textit{N} increases, and the model converges faster with fewer clients. This suggests that the greater the number of clients, the higher the data heterogeneity, which is easy to explain: when the total amount of data is fixed, a larger \textit{N} implies that each client receives less data, and with the \textit{Dir($\beta$)} partitioning, the heterogeneity becomes more severe. According to Fig. \ref{fig1} (b), (c), (d), when \textit{C} is low (a small portion of clients participate in training), the model convergence process exhibits significant oscillations under different levels of heterogeneity, especially when $\beta=0.1$. This is because, compared to a higher \textit{C}, the model differences between client groups selected at different rounds with a low sampling ratio become larger. Therefore, increasing the sampling ratio \textit{C} can significantly improve this issue. Both FedAvg and MOON show more stable convergence and higher accuracy when $C=1.0$. On the other hand, increasing the sampling ratio will raise the communication overhead during the entire training process of the model. McMahan et al. \cite{mcmahan2017communication} found through experiments that the number of communication rounds of the model when $C=0.1$ is 2 to 3 times that when $C=1.0$. However, from the perspective of the entire training process of the model, the amount of parameter exchange of the latter is 3 to 5 times that of the former, and the probability of the occurrence of stragglers during the training process increases. In addition, in some scenarios, when the client sampling ratio exceeds a certain threshold, adding more clients will bring negative effects. Thus, a scenario with a large number of clients and a low sampling ratio presents a more challenging federated learning situation.\\
Based on this, we propose a knowledge distillation with teacher-student inequitable aggregation method, which leverages the knowledge of all clients even when only a small subset of clients participate in training, balancing both model performance and training efficiency.
\begin{figure}
    \centering
    \includegraphics[width=1.0\linewidth]{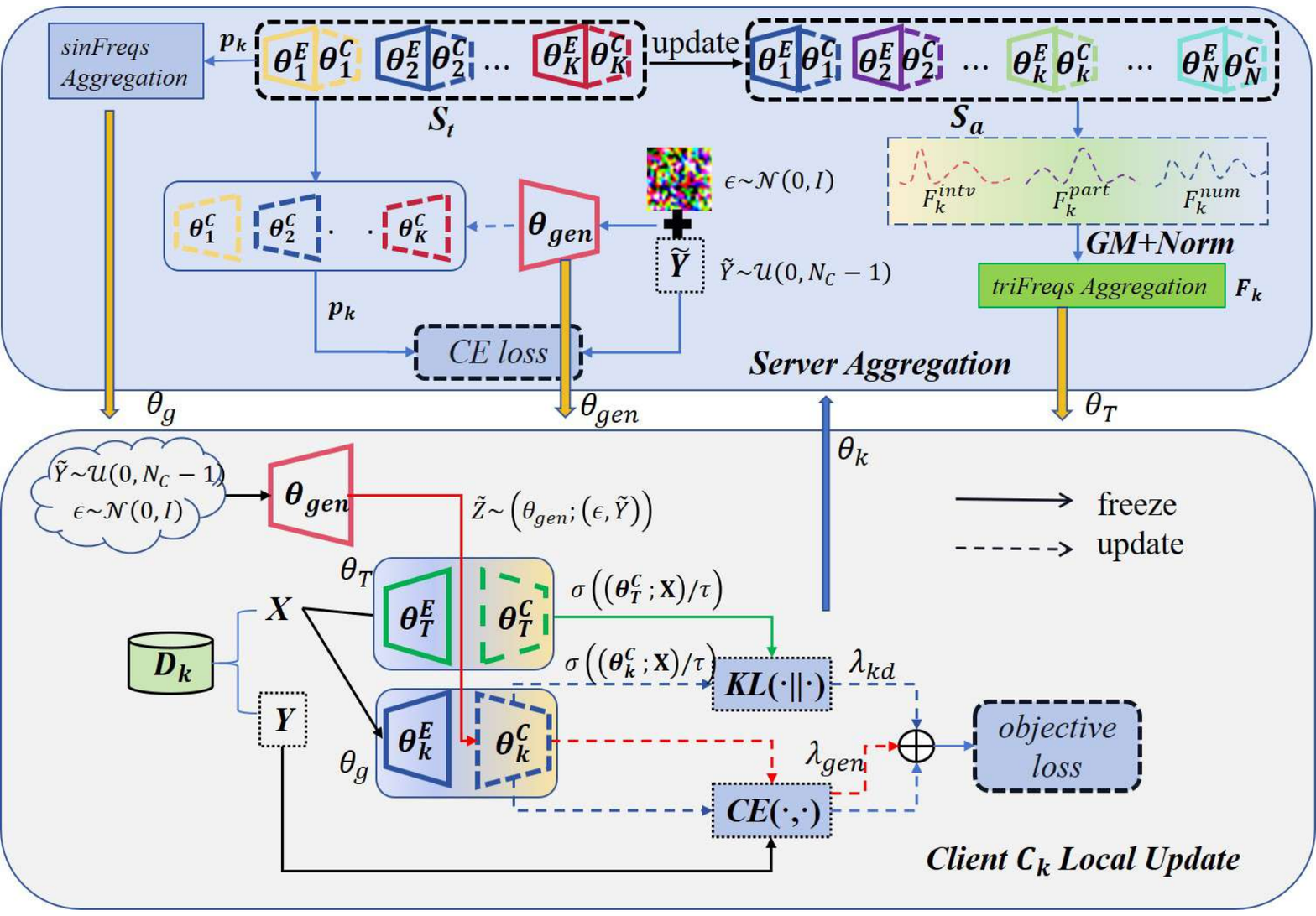}
    \caption{The figure shows the overall architecture of KDIA, where the white section represents local updates and the blue section represents server-side aggregation. The generator $\bm{\theta}_{gen}$ is trained on the server and then distributed to the clients, with the noise and label sampling methods being the same.}
    \label{fig2}
\end{figure}
\begin{figure*}
    \centering
    \includegraphics[width=\linewidth]{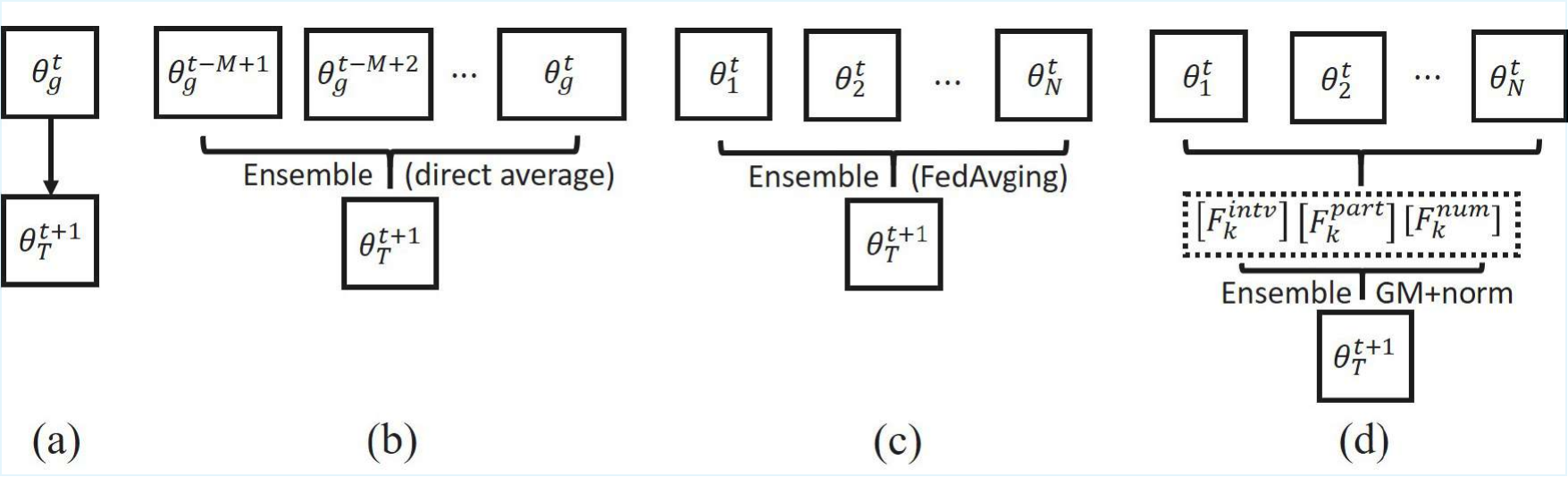}
    \caption{The figure illustrates the different ways in which the teacher model is acquired in knowledge distillation.}
    \label{fig3}
\end{figure*}
\section{Method}\label{sec3}
This section provides a detailed explanation of KDIA, with Section \ref{ss3} and \ref{ss4} covering inequitable aggregation and self-knowledge distillation, respectively, Section \ref{ss5} discussing data generation to assist training, and Section \ref{ss6} presenting the analysis of KDIA. The overall framework of KDIA is shown in Fig. \ref{fig2}, where the blue parts represent server-side aggregation (the student model aggregates the clients sampled in each round, while the teacher model performs a weighted aggregation of all clients based on three frequencies: participation interval, participation counts, and data volume proportions; conditional generator training), and the white parts represent local client training (self-knowledge distillation is performed, allowing us to choose the final model between the teacher and student; generate features with approximately uniform class distribution locally to assist in training).
\subsection{Inequitable Aggregation}\label{ss3}
The existing federated learning self-distillation methods for obtaining the teacher model include: (1) using the global (student) model from the previous round as the teacher model \cite{furlanello2018born,yang2019snapshot,kim2020self} (Fig. \ref{fig3} (a)); (2) aggregating the global models from the previous \textit{M} rounds to form the teacher model \cite{yao2021local} (Fig. \ref{fig3} (b)); (3) directly use the client data volume to conduct a weighted average for all clients \cite{zhou2022handling} (Fig. \ref{fig3} (c)). Method (1) aggregates and updates based on the clients sampled in each round, but when only a few clients participate in training, it is prone to forgetting previous knowledge, leading to limited effectiveness. Method (2) has difficulty determining the aggregation weights of models from the previous \textit{M} rounds, and when certain clients are excluded from training for a long time, their knowledge cannot be incorporated. Method (3) assigns excessive weight to clients with large amounts of data that did not participate in training, which can prevent other participating clients from making their due contribution. In the first two methods, the teacher model is updated without utilizing all the clients, leading to issues such as inability to guide due to slow updates and knowledge forgetting due to overly rapid updates.\\
To address the shortcomings of the teacher model update methods mentioned above and enable it to effectively capture the knowledge of all clients, we propose aggregating all client models to obtain the teacher model. for this, two issues need to be considered:
\begin{itemize}
    \item[(1)] How to evaluate the contribution of clients that did not participate in training?
    \item[(2)] How to calculate the weights for each client?
\end{itemize}
In view of the problems existing in Method (3) when aggregating the teacher model, we consider three factors: client participation intervals, participation counts, and data volume proportions, and propose the \textit{triFreqs} weighting method (Fig. \ref{fig3} (d)). Assuming the teacher model is $\langle\bm{\theta}_T^E,\bm{\theta}_T^C\rangle$ and the student model is $\langle\bm{\theta}_k^E,\bm{\theta}_k^C\rangle$, the \textit{triFreqs} weighting method for round \textit{t} is as follows:
\begin{itemize}
    \item[(1)] \textbf{participation intervals frequency $F_k^{intv}$}
\end{itemize}
\begin{equation}\label{eq5}
    F_k^{intv}=\frac{\exp^{-(t-t_k)}}{\sum_{k\in S_a}\exp^{-(t-t_k)}},
\end{equation}
where $t_k=
\begin{cases}
\textit{t}, \text{if k is selected}, \\
\textit{not change}, \text{otherwise}. & 
\end{cases}$ Let $t_k$ be initialized to -1. $F_k^{intv}$ will assign smaller weights to clients that have not participated in training for a long time. If client \textit{k} participates in the current training and $t-t_k=0$, then:
\begin{equation}\label{eq6}
    \begin{split}
        0<F_k^{int\nu}&=\frac{\exp^{-(t-t_k)}}{\sum_{k\in S_a}\exp^{-(t-t_k)}}\\
        &=\frac{1}{K+\sum_{k\in\{S_a-S_t\}}\exp^{-(t-t_k)}}\\
        &=\frac{1}{K+(N-K)\exp^{-t-1}}<\frac{1}{K}.
    \end{split}
\end{equation}
If client \textit{k} has not participated in training, since $t-t_k>0$, this client will have a smaller proportion. Therefore, for all clients, $0<F_k^{intv}<1/K$.
\begin{itemize}
    \item[(2)] \textbf{participation counts frequency $F_k^{part}$}
\end{itemize}
\begin{equation}\label{eq7}
    F_k^{part}=\frac{N_k^{part}}{\sum_{k\in S_a}N_k^{part}},
\end{equation}
where $N_k^{part}$ is the number of times client \textit{k} has participated in past rounds, initially set to 0. If client \textit{k} participates in round \textit{t}, $N_k^{part}$ is incremented by 1; otherwise, it remains unchanged. $F_k^{part}$ assigns smaller weights to clients with fewer participation counts, and $0\leq{F_k^{part}}\leq1/K$. When client \textit{k} participates in every round, the equality on the right side holds. As the number of training rounds increases, the participation counts $N_k^{part}$ of each client tend to become similar, and $F_k^{part}$ becomes approximately uniformly distributed, as shown in Fig. \ref{fig5} (b).
\begin{itemize}
    \item[(3)] \textbf{data volume proportions $F_k^{num}$}
\end{itemize}
\begin{equation}\label{eq8}
    F_k^{num}=\frac{N_k^{num}}{\sum_{k\in S_a}N_k^{num}}.
\end{equation}
Eq. (\ref{eq8}) represents the version of $p_k$ for all clients in Eq. (\ref{eq1}), where $F_k^{num}$ depends on the data partitioning method. After obtaining the three frequency factors mentioned above, their geometric mean is used for aggregation to obtain $F_k^{tri}$, which is then normalized to get the weight factor $F_k$ required for the teacher model aggregation.
\begin{equation}\label{eq9}
    F_k=\frac{F_k^{tri}}{\sum_{k\in S_a}F_k^{tri}},
\end{equation}
where $0\leq F_{k}^{tri}=\sqrt[3]{F_{k}^{intv}\cdot F_{k}^{part}\cdot F_{k}^{num}}<\sqrt[3]{\frac{1}{K^{2}}\cdot F_{k}^{num}}$. In Eq. (\ref{eq9}), the geometric mean is used to calculate the current \textit{triFreqs} weight of the client. This choice is made because the geometric mean is less influenced by extreme values (such as zero values) compared to the arithmetic or harmonic mean (as discussed in Section \ref{ss8}, where different averaging methods were experimentally tested), and it also helps to avoid large oscillations during the update process.\\
Let $\langle\bm{\theta}_T^E,\bm{\theta}_T^C\rangle$ be the aggregation result of all clients using $F_k$, and $\langle\bm{\theta}_g^E,\bm{\theta}_g^C\rangle$ be the aggregation result of the selected client set $S_t$. For convenience, we refer to the aggregation methods for the student models as \textit{sinFreqs} (for a single frequency). Thus:
\begin{equation}\label{eq10}
    \begin{split}
        \langle\bm{\theta}_T^E,\bm{\theta}_T^C\rangle=\sum_{k\in S_a}F_k\cdot\langle\bm{\theta}_k^E,\bm{\theta}_k^C\rangle,\\
        \langle\bm{\theta}_g^E,\bm{\theta}_g^C\rangle=\sum_{k\in S_t}p_k\cdot\langle\bm{\theta}_k^E,\bm{\theta}_k^C\rangle.
    \end{split}
\end{equation}
In Eq. (\ref{eq10}), if client \textit{k} participates in the training in the \textit{t}-th round, the teacher's $\langle\bm{\theta}_k^E,\bm{\theta}_k^C\rangle$ will also be updated accordingly (this is the process through which the teacher acquires new knowledge); otherwise, they remain unchanged.\\
In inequitable aggregation, only a subset of clients participates in the aggregation for the student model, while all clients are involved in the aggregation for the teacher model. During training, both the teacher and student models are consistently updated using this approach, but the final target model is selected based on superior performance. The sampling ratio \textit{C} plays a crucial role in influencing the results. As \textit{C} increases, the \textit{tirFreqs} weight calculation reverts to the FedAvg method. Therefore, this approach is more suitable for federated learning scenarios where \textit{N} is large and \textit{C} is low.
\subsection{Self-Knowledge Distillation}\label{ss4}
for the sample set $\bm{D}_k$, in our method, self-distillation and response-based knowledge distillation are employed, and the knowledge distillation process is generally performed using \text{KL} divergence or cross-entropy loss:
\begin{equation}\label{eq11}
    \begin{split}
        l_{kd}(\langle\bm{\theta}_{k}^{E},\bm{\theta}_{k}^{C}\rangle)
    & =\frac{\lambda_{kd}}{N_{b}}\sum_{\mathrm{\bm{D}_{b}\subseteq \bm{D}_{k}}}\mathcal{L}_{KL}(\bm{P}||\bm{Q})\\
    & =\frac{\lambda_{kd}}{N_{b}}\sum_{\mathrm{\bm{D}_{b}\subseteq \bm{D}_{k}}}(\bm{P}\mathrm{log}\bm{P}-\bm{P}\mathrm{log}\bm{Q})\\
    & =\frac{\lambda_{kd}}{N_{b}}\sum_{\mathrm{\bm{D}_{b}\subseteq \bm{D}_{k}}}(-\bm{P}\mathrm{log}\bm{Q})+C^{\prime}\\
    & =\frac{\lambda_{kd}}{N_{b}}\sum_{\mathrm{\bm{D}_{b}\subseteq \bm{D}_{k}}}\mathcal{L}_{CE}(\bm{Q},\bm{P})+C^{\prime},
    \end{split}
\end{equation}
where $\bm{P}=\sigma((\langle\bm{\theta}_T^E,\bm{\theta}_T^C\rangle;\mathrm{\bm{D}_b})/\tau)$, $\bm{Q}=\sigma((\langle\bm{\theta}_k^E,\bm{\theta}_k^C\rangle;\mathrm{\bm{D}_b})/\tau)$, $\tau$ is the distillation temperature, which primarily serves to smooth the prediction distribution and prevent gradient vanishing. It is set to 2.0 by default. $C^{\prime}$ is a constant. In Eq. (\ref{eq11}), $\lambda_{kd}$ is a hyperparameter, with a default value of 0.5 (refer to Section \ref{ss8} for parameter selection). $N_b$ is the batch size, and $\bm{D}_b$ is a set of the batch. The above equations indicate that using \text{KL} divergence and cross-entropy loss for loss computation is equivalent \cite{yuan2021revisitingknowledgedistillationlabel}.\\
During the training process, the teacher and student models mutually enhance each other. The new knowledge acquired by the teacher model during aggregation is brought by the clients currently participating in training, while historical knowledge is retained by other clients. The student model is guided by the teacher model during local training, iterating continuously to obtain a better model. Experimental results show that both the teacher and student models progress together, with the teacher model consistently outperforming the student model. The student also partially takes on the role of the teacher, achieving the goal of self-distillation.
\subsection{Generate Auxiliary Data with CG}\label{ss5}
To alleviate the heterogeneity of local data, we use the data generation approach proposed by FedGen \cite{zhu2021data} to assist local training. On the server side, unlike FedGen, which generates labels $\bm{\tilde{Y}}$ based on batch size, our method first calculates the total number of labels required, then generates the labels $\bm{\tilde{y}}$ in advance. Afterward, we sample $\bm{\epsilon}\sim\mathcal{N}(0,I)$ and $\bm{\tilde{Y}}\subseteq\bm{\tilde{y}}$ based on the batch size to generate data for training the generator $\bm{\theta}_{gen}$. The purpose of this approach is as follows:
\begin{itemize}
    \item[(1)] When generating data with a uniform distribution, the longer the data sequence, the closer the generated labels $\bm{\tilde{Y}}$ are to a uniform distribution, achieving the goal of generating IID data features. However, when generating data features using a batch size (e.g., 32, 64, 128), the independence between batches makes it difficult for $\bm{\tilde{Y}}$ to approach a uniform distribution.
    \item[(2)] Before each local round, the labels $\bm{\tilde{y}}$ can be shuffled. This introduces variability in the data generated for each batch, improving the quality of the generated data fatures and the model's generalization ability.
    \item[(3)] for clients with a data size $N_k$ smaller than the batch size, the amount of generated data features is increased to $N_k\times{E}$, where \textit{E} is the local epoch. This way, different generated samples participate in training during each local epoch.
\end{itemize}
When generating auxiliary training data $\tilde{\bm{D}_k}$ locally, the process follows the same method as described above, except that the generator network $\bm{\theta}_{gen}$ is frozen. The loss for this part is calculated as follows:
{\small
\begin{equation}\label{eq12}
    l_{gen}(\langle\bm{\theta}_k^E,\bm{\theta}_k^C\rangle)=\frac{\lambda_{gen}}{N_b}\sum_{\widetilde{\mathcal{\bm{D}}_b}\subseteq\widetilde{\mathcal{\bm{D}}_k},\widetilde{Y}\subseteq\widetilde{y}}\mathcal{L}_{CE}\left(\sigma\left(\left(\langle\bm{\theta}_k^E,\bm{\theta}_k^C\rangle;\widetilde{\mathcal{\bm{D}}_b}\right)\right),\bm{\tilde{Y}}\right),
\end{equation}
}
where $\lambda_{gen}$ is a hyperparameter, and its settings can be found in Section \ref{ss8}.\\
The final objective function is:
{\small
\begin{equation}\label{eq13}
\min_{\langle\bm{\theta}_k^E,\bm{\theta}_k^C\rangle}\mathcal{L}\left(\langle\bm{\theta}_k^E,\bm{\theta}_k^C\rangle\right)=l_{CE}(\langle\bm{\theta}_k^E,\bm{\theta}_k^C\rangle)+l_{kd}(\langle\bm{\theta}_k^E,\bm{\theta}_k^C\rangle)+l_{gen}(\langle\bm{\theta}_k^E,\bm{\theta}_k^C\rangle),
\end{equation}
}
where $l_{CE}$ is the original cross-entropy classification loss.

\begin{algorithm}
	\caption{KDIA (Server Aggregation)}
	\label{alg1}
    \begin{algorithmic}[1]
        \Require initial global model $\bm{\theta}_g^0$, teacher model    $\bm{\theta}_T^0$, generator $\bm{\theta}_{gen}^0$, teacher's local model set $\left[\bm{\theta}_k\right]_{k=1}^N$, communication round \textit{T}, generator training epoch $T_{gen}$, number of batch $B_{gen}$, and batch size $N_b$, total client number \textit{N}, sampling ratio \textit{C}
        \Ensure final model $\bm{\theta}_g$ or $\bm{\theta}_T$ (selected by performance)
            \For{$t\gets0$ to $T-1$}
		\State {Randomly select $N\times{C}$ clients from $S_a$ as a set $S_t$} 
                \For{client $k\in{S_t}$}
                        \State {send $\bm{\theta}_g^t,\bm{\theta}_T^t,\bm{\theta}_{gen}^t$ \text{to client} $k$}
                        \State {$\bm{\theta}_k^{t+1}\gets$ \textbf{LocalUpdate}($k,\bm{\theta}_g^t,\bm{\theta}_T^t,\bm{\theta}_{gen}^t$)}
                \EndFor
                \State {calculate $p_k$ and $F_k$ according Eq. (\ref{eq1}) and (\ref{eq9}})
                \State {$\left[\bm{\theta}_k\right]_{k=1}^N\gets\left[\bm{\theta}_k^{t+1}\right]_{k\in{S_t}}$} 
                \State {$\bm{\theta}_g^{t+1}\gets\sum_{k\in{S_t}}p_k\cdot\bm{\theta}_k^{t+1}$}
                \State {$\bm{\theta}_T^{t+1}\gets\sum_{k\in{S_a}}F_k\cdot\bm{\theta}_k$}
                \State {$\bm{\tilde{y}}\gets$ (sample $T_{gen}\times{B_{gen}}$ $\bm{\tilde{y}}{\sim}U(0,N_c-1)$)}
                \For{$t\gets0$ to $T_{gen}-1$}
                    \State {shuffle $\bm{\tilde{y}}$}
                    \For{each batch $N_b$ from $\left[T_{gen}\times{B_{gen}}\right]$}
                        \State {$\left[\bm{\epsilon},\bm{\tilde{Y}}\right]\gets$ (sample a batch of $\bm{\epsilon}{\sim}\mathcal{N}(0,I)$ and $\bm{\tilde{Y}}\subseteq{\bm{\tilde{y}}}$)}
                        \State {$\bm{\theta}_{gen}^{t+1}\gets$ CG training \bigg($\bm{\theta}_{gen}^t,\bm{\theta}_k^{t+1},\left[\bm{\epsilon},\bm{\tilde{Y}}\right]$\bigg) according Eq. (\ref{eq3}) and (\ref{eq4})}
                    \EndFor
                \EndFor
            \EndFor
    \end{algorithmic}
\end{algorithm}
\begin{algorithm}
	\caption{KDIA (LocalUpdate)}
	\label{alg2}
    \begin{algorithmic}[1]
        \Require global model $\bm{\theta}_g^t$, teacher model    $\bm{\theta}_T^t$, generator $\bm{\theta}_{gen}^t$, local epoch \textit{E}, batch size $N_b$, learning rate $\eta$, hyperparameters $\lambda_{kd}$ and $\lambda_{gen}$
        \Ensure $\bm{\theta}_k^{t+1}$
            \State {$\bm{\theta}_k^t\gets\bm{\theta}_g^t$}
            \If{$N_k<N_b$}
                \State $N_k=E\times{N_k}$ 
            \EndIf
            \State {$\bm{\tilde{y}}\gets$ (sample $N_k$ $\bm{\tilde{y}}{\sim}U(0,N_c-1)$)}
            \For{$t\gets0$ to $E-1$}
                \State {$\left[\bm{\epsilon},\bm{\tilde{Y}}\right]\gets$ (sample a batch of $\bm{\epsilon}{\sim}\mathcal{N}(0,I)$ and $\bm{\tilde{Y}}\subseteq{\bm{\tilde{y}}}$)}
                \State {$\tilde{\bm{D}_b}\gets$ $\Bigg(\bm{\theta}_{gen}^t;\bigg(\left[\bm{\epsilon},\bm{\tilde{Y}}\right]\bigg)\Bigg)$ }
		      \For{each batch $\bm{D}_b$ samples from $\bm{D}_k$        and $\tilde{\bm{D}_b}$}
                    \State {$\mathcal{L}(\bm{\theta}_k^t)\gets$ compute loss according Eq. (\ref{eq11}), (\ref{eq12}), and (\ref{eq13})}
                    \State {$\bm{\theta}_k^{t+1}\gets\bm{\theta}_k^t-\eta\cdot\frac{\partial\mathcal{L}(\bm{\theta}_k^t)}{\partial\bm{\theta}_k^t}$ }
                \EndFor
            \EndFor
            \State \Return $\bm{\theta}_k^{t+1}$
    \end{algorithmic}
\end{algorithm}
\subsection{Convergence Analysis}\label{ss6}
The proposed KDIA method updates $\bm{\theta}_g$ in the \textit{t}-th round as shown in Algorithm \ref{alg1} and \ref{alg2}. For convenience of expression, let $\bm{\theta}_k^t=\langle\bm{\theta}_k^E,\bm{\theta}_k^C\rangle$ and $\bm{\theta}_T^t=\langle\bm{\theta}_T^E,\bm{\theta}_T^C\rangle$ represent the model weights of client \textit{k} and the teacher at round \textit{t}, respectively. Since both the original data and generated data use cross-entropy loss, let $\mathcal{L}_{CE}(\bm{\theta}_k^t)=l_{cls} (\bm{\theta}_k^t)+l_{gen}(\bm{\theta}_k^t)$ and $\mathcal{L}_{KL}(\bm{\theta}_k^t;\bm{\theta}_T^t)=l_{kd}(\bm{\theta}_k^t;\bm{\theta}_T^t)$. The optimization problem for client \textit{k} at round \textit{t} can then be formulated as:
{\small
\begin{equation}\label{eq14}
\begin{split}
    \bm{\theta}_{k}^{t+1}&\approx\underset{\bm{\theta}_{k}^{t}}{\operatorname*{\operatorname*{argmin}}}\{\mathcal{L}_{CE}(\bm{\theta}_{k}^{t})+\mathcal{L}_{KL}(\bm{\theta}_{k}^{t};\bm{\theta}_{T}^{t})\}\\
        &=\underset{\bm{\theta}_{k}^{t}}{\operatorname*{\operatorname*{argmin}}}\left\{\mathcal{L}_{CE}(\bm{\theta}_{k}^{t})+\frac{\mu}{2N_{k}}\sum_{j=1}^{N_{k}}\text{KL}\left(\sigma{\left((\bm{\theta}_{T}^{t},X_{j})\right)}||\sigma{\left((\bm{\theta}_{k}^{t},X_{j})\right)}\right)\right\},
\end{split}
\end{equation}
}
where $\mu>0$.\\
\textbf{Assumption 1.}
\begin{itemize}
    \item[1.] \textit{(Lower bound on eigenvalue) for all clients k, $\mathcal{L}_{CE}^k(\bm{\theta}_k)$ is L-smooth, meaning that for all , it satisfies $\left\|\nabla\mathcal{L}^k(\bm{\theta}_k)-\nabla\mathcal{L}^k(\bm{\theta}_k^{\prime})\right\|\leq L\|\bm{\theta}_k-\bm{\theta}_k^{\prime}\|$. Additionally, assume that the minimum eigenvalue of the Hessian matrix of the client loss function $\nabla^2\mathcal{L}^k(\bm{\theta}_k)$ has a lower bound $\lambda_{min}$}.
    \item[2.] \textit{(Bounded dissimilarity) for all clients k and any $\bm{\theta}_k$, it holds that\\ $\mathbb{E}_k\left[\|\nabla\mathcal{L}^k(\bm{\theta}_k)\|^2\right]\leq B^2\|\nabla\mathcal{L}(\bm{\theta}_k)\|^2$}.
    \item[3.] \textit{for all clients k, the softmax activation function outputs a probability vector, i.e., $[\sigma_k(\cdot)]\geq{0}$, and for class $j\in[N_{c}]$, $\sum_{j=1}^{C}[\sigma_{k}(\cdot)]_{j}=1$. Assume that for any sample X, parameters $\bm{\theta}$, and client k, there exists a constant $\delta>0$ such that $\min_{j\in\{1,\cdots,C\}}[\sigma_k(\cdot)]\geq\delta>0$, and there exists a constant $L_h>0$ such that $\|\sigma_k(\bm{\theta},X)-\sigma_k(\bm{\theta}^{\prime},X)\|\leq L_h\|\bm{\theta}-\bm{\theta}^{\prime}\|$}.
\end{itemize}
\textbf{Assumption 2.}
\begin{itemize}
    \item[1.] \textit{($\gamma$-inexact solution) for a function $\mathcal{L}(\bm{\theta}^{\prime},\bm{\theta}_k)=\mathcal{L}(\bm{\theta}_k)+\frac{\mu}{2}\|\bm{\theta}_k-\bm{\theta}^{\prime}\|^2$ and $\gamma\in[0,1]$, if $\|\mathcal{L}(\bm{\theta}^*)+\mu(\bm{\theta}^*-\bm{\theta}_k)\|\leq\gamma\|\mathcal{L}(\bm{\theta}_k,\bm{\theta}_k)\|$, then $\bm{\theta}^*$ is a $\gamma$-inexact solution for the optimization problem $\min_{\bm{\theta}_k}\mathcal{L}(\bm{\theta},\bm{\theta}_k)$.Therefore, for the Eq. (\ref{eq14}), we have:}.
\end{itemize}
\begin{equation*}
    \left\|\nabla\mathcal{L}_{CE}^{k}(\bm{\theta}_{k}^{t+1})+\frac{\mu L_{h}}{\delta}(\bm{\theta}_{k}^{t}-\bm{\theta}_{T}^{t})\right\|\leq\gamma\|\nabla\mathcal{L}(\bm{\theta}_{T}^{t},\bm{\theta}_{T}^{t})\|,
\end{equation*}
where $\gamma\in{[0,1)}$. \\
We make the same assumptions 1.1, 1.2, and 2.1 as in FedProx \cite{li2020federated}, and assumption 1.3 as in FedGKD \cite{yao2021local}.\\
\textbf{Theorem 1.} (\textit{Convergence}) Under assumptions 1 and 2, for each round \textit{t} of selected client set $S_t$, i.e., the \textit{K} clients, if the selected $\mu,\gamma$, and \textit{K} satisfy:
{\footnotesize
\begin{align*}
            \varphi:&=\frac{\mu L_h}{\delta}+\lambda_{min}>0,\\
\rho & =\frac{\delta}{\mu L_{h}}\left(1-\gamma\mathrm{B}-\frac{LB(1+\gamma)}{\varphi}\right) \\
 & -\frac{1}{\varphi}\left(\frac{\sqrt{2}B(1+\gamma)}{\sqrt{K}}+\frac{L(1+\gamma)^{2}B^{2}}{2\varphi}+\frac{\left(2\sqrt{2K}+2\right)LB^{2}(1+\gamma)^{2}}{\varphi K}\right)>0.
\end{align*}
}
After \textit{T} rounds of training, we have:
\begin{equation*}
    \mathbb{E}_{S_t}[\mathcal{L}(\bm{\theta}^{t+1})]\leq\mathcal{L}(\bm{\theta}^t)-\rho\|\nabla\mathcal{L}(\bm{\theta}^t)\|^2.
\end{equation*}
For the proof of \textbf{Theorem 1}, please refer to the proof of Theorem 4 in FedProx \cite{li2020federated}.
Taking the total expectation of the above inequality, considering all sources of randomness (local stochastic gradient descent, random client sampling), and scaling, we get:
\begin{equation*}
    \rho\sum_0^{T-1}\|\mathcal{L}(\bm{\theta}^t)\|^2\leq \mathcal{L}(\bm{\theta}^0)-\mathcal{L}(\bm{\theta}^*).
\end{equation*}
Dividing both sides of the inequality by \textit{T}, we get:
\begin{equation*}
    \min_{t\in[T]}\mathbb{E}[\nabla\mathcal{L}(\bm{\theta}^t)]\leq\frac{1}{T}\sum_{0}^{T-1}\|\nabla\mathcal{L}(\bm{\theta}^t)\|^2\leq\frac{\mathcal{L}(\bm{\theta}^0)-\mathcal{L}(\bm{\theta}^*)}{\rho T}.
\end{equation*}
The final convergence rate obtained is $O(1/T)$.
\section{Experiments}\label{s5}
In this section, we present comparative experiments on KDIA, including comparisons of test accuracy and convergence curves with baseline methods, ablation studies, and parameter selection.
\subsection{Experimental Setup}\label{ss7}
\textbf{Datasets.} We evaluate the performance of various methods on three image classification tasks: CIFAR-10/100 and CINIC-10. CIFAR-10 and CINIC-10 have 10 classes each, while CIFAR-100 has 100 classes. The input image size for all three datasets is 32×32.
\begin{figure*}
        \begin{minipage}{0.32\linewidth}
		\vspace{3pt}
		\centerline{\includegraphics[width=\textwidth]{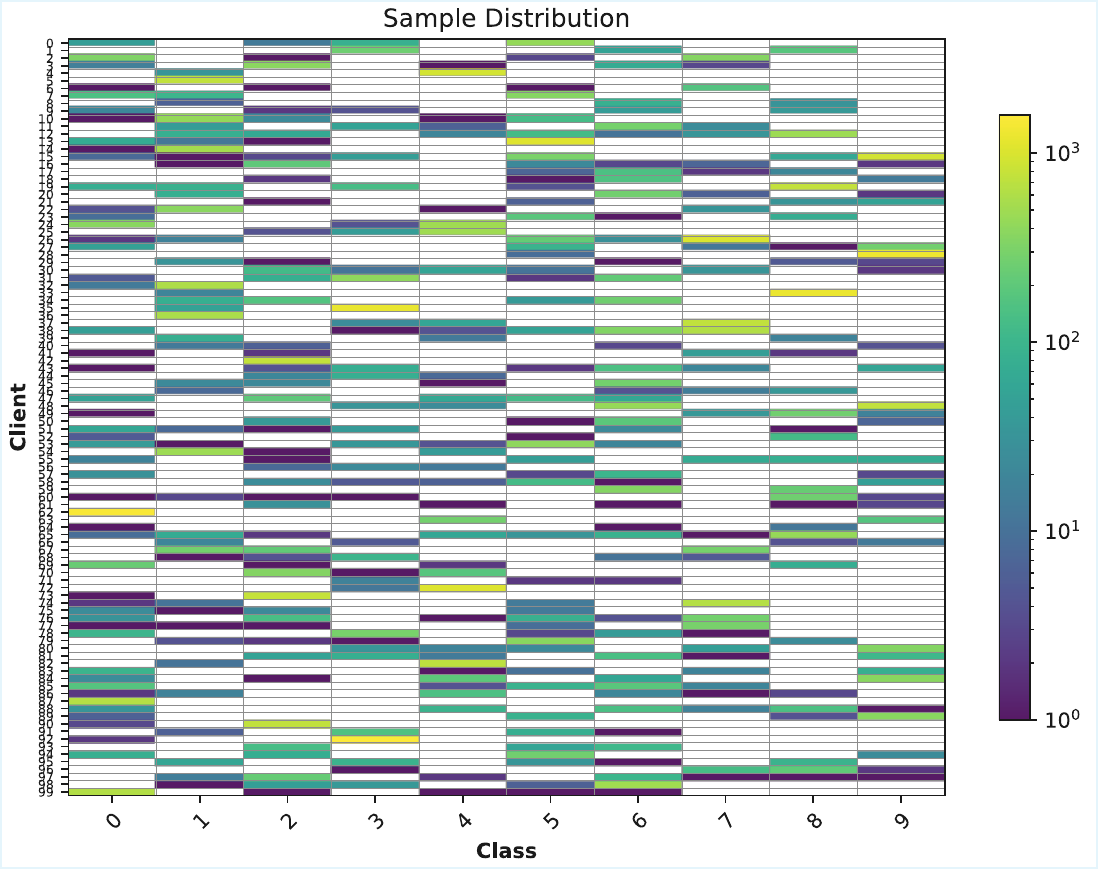}}
		\centerline{\footnotesize (a) $\beta=0.1$}
	\end{minipage}
	\begin{minipage}{0.32\linewidth}
		\vspace{3pt}
		\centerline{\includegraphics[width=\textwidth]{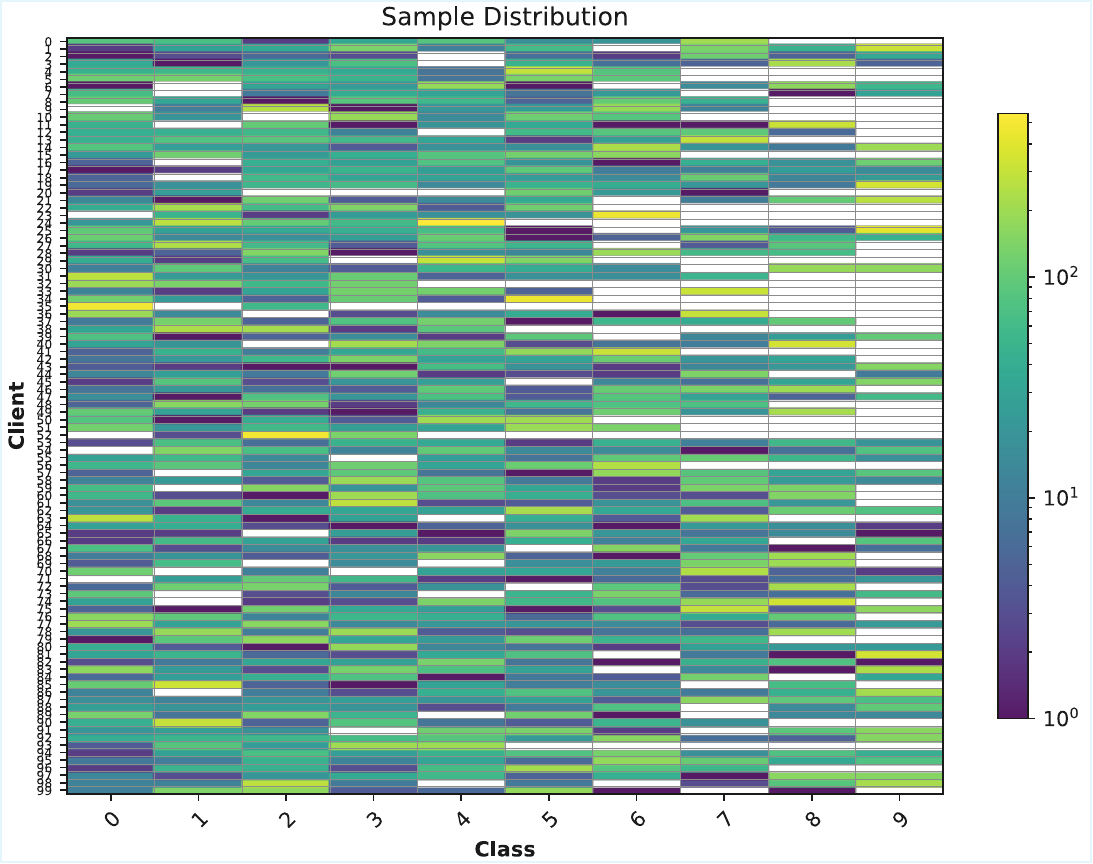}}
		\centerline{\footnotesize (b) $\beta=0.5$}
	\end{minipage}
    	\begin{minipage}{0.32\linewidth}
		\vspace{3pt}
		\centerline{\includegraphics[width=\textwidth]{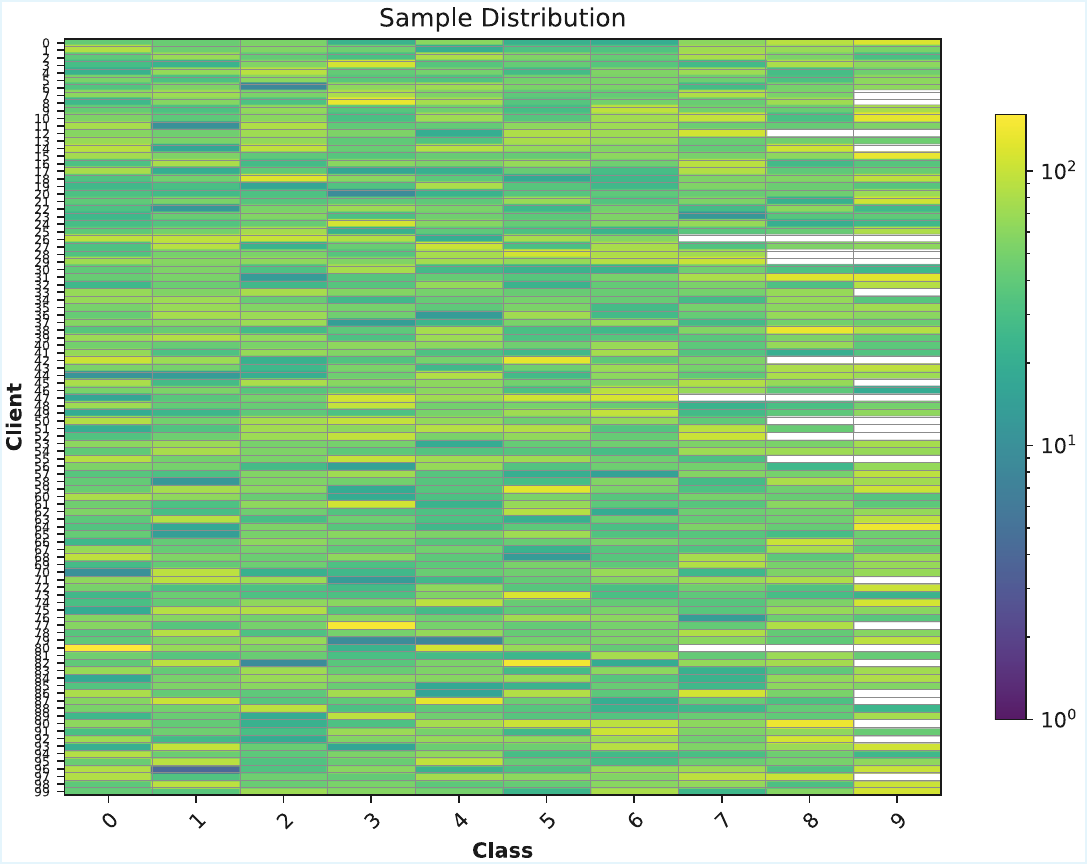}}
		\centerline{\footnotesize (c) $\beta=5.0$}
	\end{minipage}
	\caption{Distribution of data under different \textit{Dir($\beta$)} partitions on the CIFAR-10 dataset. The x-axis represents the class index, the y-axis represents the client index, and the color bar indicates the sample size.}
	\label{fig4}
\end{figure*}

\textbf{Heterogeneity setup.} We follow the same setup as MOON \cite{li2021model}, where data is partitioned using \textit{Dir($\beta$)}, with $\beta$ taking values of 0.1, 0.5, and 5.0. A lower $\beta$ indicates a higher degree of data heterogeneity. The data distribution is shown in Fig. \ref{fig4}. Additionally, \textit{N} is set to 100 by default, and \textit{C} is set to 0.1, which exacerbates data heterogeneity to some extent.

\textbf{Implementation details.} For all methods, the default number of local epochs per client is 10, and the batch size is 64. We use SGD as the optimizer, with a learning rate of 0.01, weight decay of 0.00001, and momentum of 0.9. For the three datasets, we use the same simple CNN as the feature extractor and classifier as in MOON \cite{li2021model}. The output feature dimension of the extractor is: $16\times5\times5$ ($C\times{H}\times{W}$). The global number of epochs is set to 200 by default. The generator settings are the same as in FedCG \cite{wu2021fedcg}, with 10 training epochs, a batch size of 64, and 200 batches generated. In Gaussian distribution sampling, \textit{I} is set to 100, and the Adam optimizer is used with a learning rate of 0.001 and weight decay of 0.00001.

\textbf{Baselines.} We adopt 7 methods as baselines: FedAvg \cite{mcmahan2017communication}, FedAvgM \cite{hsu1909measuring}, FedProx \cite{li2020federated}, MOON \cite{li2021model}, FedGKD \cite{yao2021local}, FedDisco \cite{ye2023feddisco} and FedCOG \cite{yefake}. For FedAvgM, the parameter $\alpha$ is selected from $\{0.7, 0.9, 0.99\}$, with the default value of 0.9. For FedProx, the parameter $\mu$ is selected from $\{0.001, 0.01, 0.1, 1.0\}$, with the default value of 0.01. For MOON, the parameter $\mu$ is selected from $\{1.0, 5.0, 10.0\}$, with the default value of 5.0. For FedGKD, the parameter $\mu$ is selected from $\{0.02, 0.1, 0.2\},$ with the default value of 0.2. For FedDisco, the parameters \textit{a} and \textit{b} are set to 0.5 and 0.1 by default. For FedCOG, the default settings of $\lambda_{dis}$ and $\lambda_{kd}$ are 0.1 and 0.01. For our method, the parameters $\lambda_{kd}/\lambda_{gen}$ are set to 0.5/0.01 when $\beta=0.1$, 0.5/1.0 when $\beta=0.5$, and 0.5/0.01 when $\beta=5.0$, respectively. All experiments are conducted on an RTX 2080ti GPU.
\subsection{Main Results}\label{ss8}
\textbf{Test accuracy.} For accuracy, KDIA and 6 baseline methods were tested on the CIFAR-10/100 and CINIC-10 datasets, with the results shown in Table \ref{tab2}. In extreme data heterogeneity conditions ($N=100/C=0.1/\beta=0.1$), KDIA achieved SOTA performance. Specifically, on CIFAR-10, the teacher model's predictions outperformed the baseline methods by 6.88\%, 4.02\%, 4.60\%, 8.47\%, 5.79\%, and 3.20\%, respectively. On CIFAR-100, KDIA outperformed the baseline methods by 2.74\%, 2.70\%, 2.54\%, 2.32\%, 2.33\%, and 2.87\%, respectively. On CINIC-10, KDIA outperformed the baseline methods by 3.70\%, 1.69\%, 3.17\%, 4.92\%, 4.36\%, and 1.81\%, respectively. When $\beta=0.1/\beta=5.0$, KDIA performed better in most cases (with improvements around 1\%-2\%). However, on the CINIC-10 dataset with $\beta=5.0$, KDIA performed worse than MOON. This could be due to the fact that CINIC-10 contains 90,000 training samples, and in near-IID conditions, the effects of \textit{N} and \textit{C} on the results are much smaller than those on CIFAR datasets.\\
Additionally, it is worth noting that on the CIFAR-100 dataset, when $\beta=5.0$, the performance of all methods was lower than when $\beta=0.5$. Theoretically, under the same settings, test results should improve as heterogeneity decreases. The anomalous result may be due to the combined effects of total client count, client sampling ratio, category number, and the heterogeneity of the partitioning. Considering two scenarios under the premise of $N=100$: (1) Each client contains data from only a few categories. (2) The 100 clients' data distribution is nearly IID. For scenario (1), although each client only contains a few categories, these categories are "\textit{sufficient in quantity}". Clients can learn adequately, and as training proceeds, all categories will be involved, ensuring data diversity. However, in scenario (2), when the distribution is close to IID, the data available to each client is "\textit{full in categories but small in quantity}" (e.g., 100 categories, 5 samples per category), making it impossible for clients to fully learn each class. As a result, multiple "\textit{mediocre}" local clients aggregate into a global model with worse performance.
\begin{table*}
    \centering
    \caption{Test accuracy of baseline methods and our method under different levels of heterogeneity on three datasets. The results are presented as mean±std from three random seed test results.}
    \label{tab2}
    \resizebox{1.0\textwidth}{!}{
    \begin{tabular}{cccccccccc}
         \toprule
            \multirow{2}{*}{\textbf{Method}} & \multicolumn{3}{c}{\textbf{CIFAR-10}} & \multicolumn{3}{c}{\textbf{CIFAR-100}} & \multicolumn{3}{c}{\textbf{CINIC-10}}\\
            & $\beta=0.1$ & $\beta=0.5$ & $\beta=5.0$ & $\beta=0.1$ & $\beta=0.5$ & $\beta=5.0$ & $\beta=0.1$ & $\beta=0.5$ & $\beta=5.0$\\
            \cline{1-10}
            FedAvg \cite{mcmahan2017communication}& $50.41\pm{1.91}$ & $64.36\pm{0.69}$ & $67.17\pm{0.82}$ & $22.18\pm{0.28}$& $22.43\pm{0.17}$ & $19.79\pm{0.30}$ & $42.37\pm{1.99}$ & $51.19\pm{0.28}$ & $53.25\pm{0.66}$\\
            FedAvgM \cite{hsu1909measuring}& $53.27\pm{1.17}$&$61.44\pm{0.17}$&	$64.98\pm{0.64}$&	$22.22\pm{0.44}$	&$22.10\pm{0.49}$&	$20.26\pm{0.17}$&	$44.38\pm{0.85}$&	$49.70\pm{0.33}$	&$52.00\pm{0.69}$\\
            FedProx \cite{li2020federated}& $52.69\pm{1.64}$&$64.15\pm{0.41}$&$67.03\pm{0.46}$&	$22.38\pm{0.30}$&$22.66\pm{0.53}$&$21.49\pm{0.40}$&$42.90\pm{1.58}$&	$51.06\pm{0.40}$&$53.17\pm{0.58}$\\
            MOON \cite{li2022federated}& $48.82\pm{2.61}$&$64.69\pm{0.10}$&	$68.24\pm{0.50}$&$22.60\pm{0.46}$&$25.62\pm{0.84}$&	$25.34\pm{0.37}$&$41.15\pm{1.47}$&\underline{$51.83\pm{0.32}$}&	\bm{$54.16\pm{0.59}$}\\
            FedGKD \cite{wang2020tackling}& $51.50\pm{1.82}$&$63.11\pm{0.22}$&	$67.31\pm{0.35}$&$22.59\pm{0.41}$&$22.94\pm{0.40}$&	$21.80\pm{0.45}$&$41.71\pm{2.75}$&$51.52\pm{0.30}$&	$53.33\pm{0.10}$\\
            FedDisco \cite{ye2023feddisco}& $54.09\pm{1.74}$&$63.43\pm{0.32}$&	$67.22\pm{0.39}$&$22.05\pm{0.66}$&$22.58\pm{0.54}$&	$21.40\pm{0.73}$&$44.26\pm{1.20}$&$51.44\pm{0.47}$&	$52.94\pm{0.20}$\\
            FedCOG \cite{yefake}& $49.54\pm{1.88}$&$62.36\pm{0.66}$&	$65.99\pm{0.69}$&$22.33\pm{0.84}$&$22.40\pm{1.12}$&	$21.21\pm{0.28}$&\underline{$44.96\pm{1.03}$}&$51.57\pm{0.51}$&	$53.78\pm{0.35}$\\
            Ours(S) & \underline{$54.11\pm{1.42}$}&\underline{$65.07\pm{0.39}$}&	\underline{$69.11\pm{0.14}$}&\underline{$23.37\pm{0.17}$}&\underline{$25.34\pm{0.54}$}&	\underline{$25.01\pm{0.46}$}&$43.45\pm{1.82}$&$51.36\pm{0.44}$&
            $53.62\pm{0.52}$\\
            Ours(T) & \bm{$57.29\pm{0.97}$}&\bm{$65.91\pm{0.76}$}&	\bm{$69.42\pm{0.10}$}&\bm{$24.92\pm{0.51}$}&\bm{$26.43\pm{0.60}$}&	\bm{$26.40\pm{0.65}$}&\bm{$46.07\pm{1.03}$}&\bm{$52.22\pm{0.37}$}&\underline{$53.99\pm{0.62}$}\\
        \bottomrule
    \end{tabular}
}
\end{table*}
\begin{table}
    \centering
    \caption{The table shows the test accuracy of the teacher models aggregated using different weight calculation methods. \textit{triFreqs(AM)/triFreqs(GM)} represent aggregation after averaging the three frequencies using arithmetic/geometric mean, respectively.}
    \label{tab3}
    \resizebox{1.0\linewidth}{!}{
        \begin{tabular}{cccc}
        \toprule
            CIFAR-10&$\beta=0.1$ & $\beta=0.5$ & $\beta=5.0$\\
            \cline{1-4}
            \textit{$F_k^{intv}$}&$55.53\pm{1.18}(54.89\pm{1.68})$	&$64.91\pm{0.85}(64.68\pm{0.91})$&$67.89\pm{0.57}(67.74\pm{1.39})$\\
            \textit{$F_k^{part}$}&$56.45\pm{1.25}(54.80\pm{1.55})$	&$65.37\pm{0.70}(65.02\pm{0.83})$&$68.36\pm{0.88}(67.76\pm{1.03})$\\
            \textit{$F_k^{num}$}&$55.69\pm{1.07}(54.96\pm{1.51})$	&$64.95\pm{0.79}(64.61\pm{0.81})$&$68.41\pm{0.97}(68.10\pm{1.10})$\\
            \textit{triFreqs(AM)}&$56.70\pm{1.31}(54.89\pm{1.57})$	&\bm{$65.74\pm{0.73}(64.75\pm{0.63})$}&$68.48\pm{1.14}(67.92\pm{1.26})$\\
            \textit{triFreqs(GM)}&\bm{$57.11\pm{1.89}(55.73\pm{1.35})$}	&$65.36\pm{0.57}(64.56\pm{0.26})$&\bm{$68.56\pm{1.17}(68.16\pm{1.24})$}\\
        \bottomrule
        \end{tabular}
    }
\end{table}
\begin{figure*}[ht]
    \begin{minipage}{0.49\linewidth}
		\vspace{3pt}
		\centerline{\includegraphics[width=\textwidth]{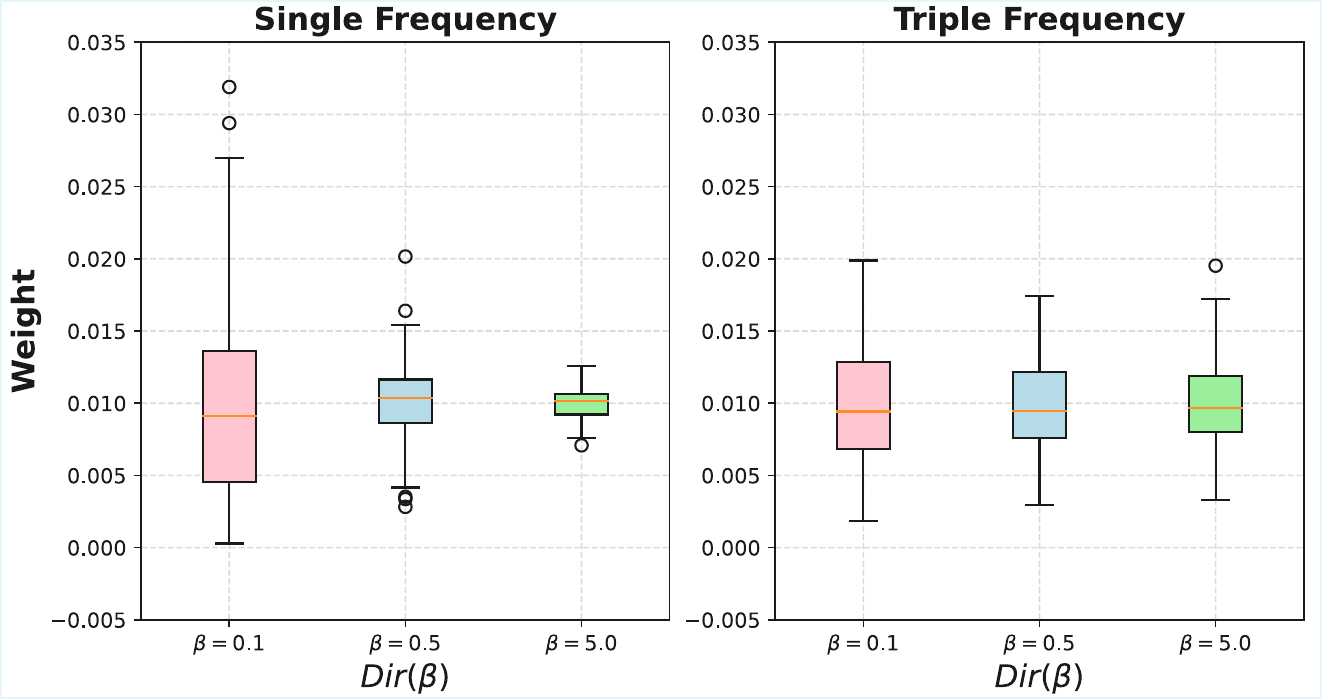}}
		\centerline{\footnotesize (a) sinFreqs/triFreqs}
	\end{minipage}
	\begin{minipage}{0.49\linewidth}
		\vspace{3pt}
		\centerline{\includegraphics[width=\textwidth]{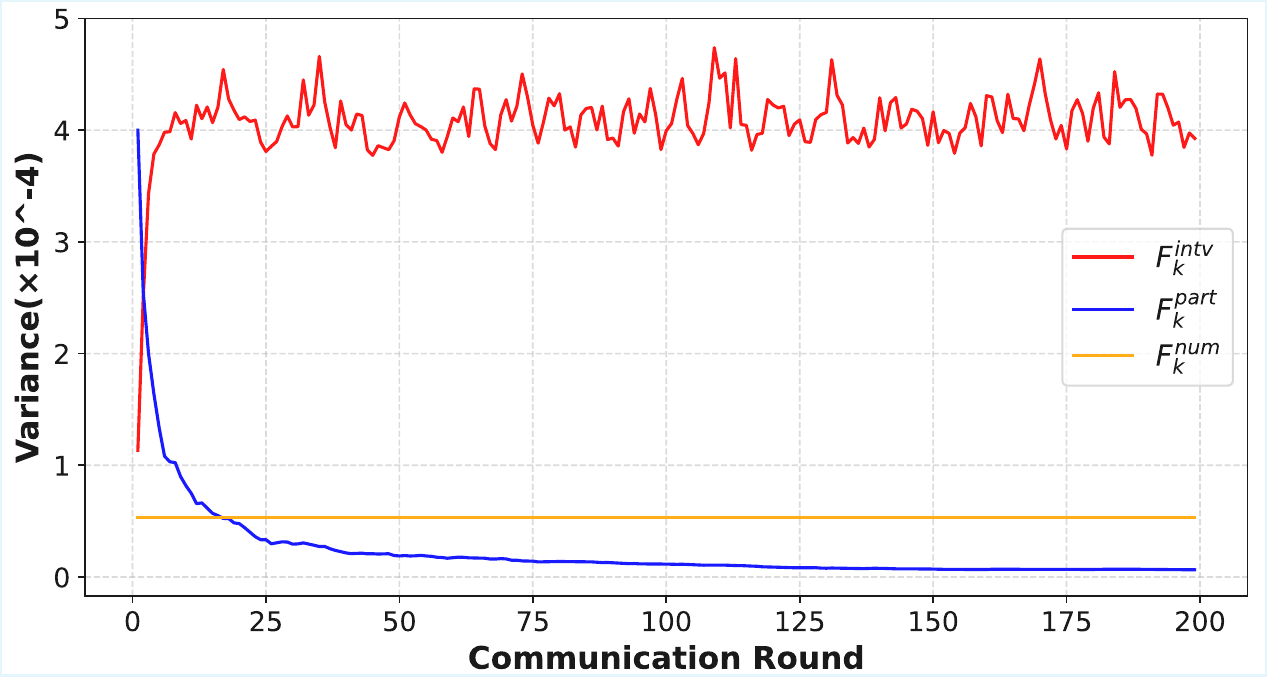}}
		\centerline{\footnotesize (b) Variance}
	\end{minipage}
	\caption{(a) shows the distribution of clients for \textit{sinFreqs} (left) / \textit{triFreqs} (right) at different levels of heterogeneity. (b) illustrates the training variation of client participation interval frequency variance, participation count frequency variance, and data proportion variance when $\beta=0.5$.}
	\label{fig5}
\end{figure*}
\begin{figure*}[ht]
       \begin{minipage}{0.32\linewidth}
		\vspace{3pt}
		\centerline{\includegraphics[width=\textwidth]{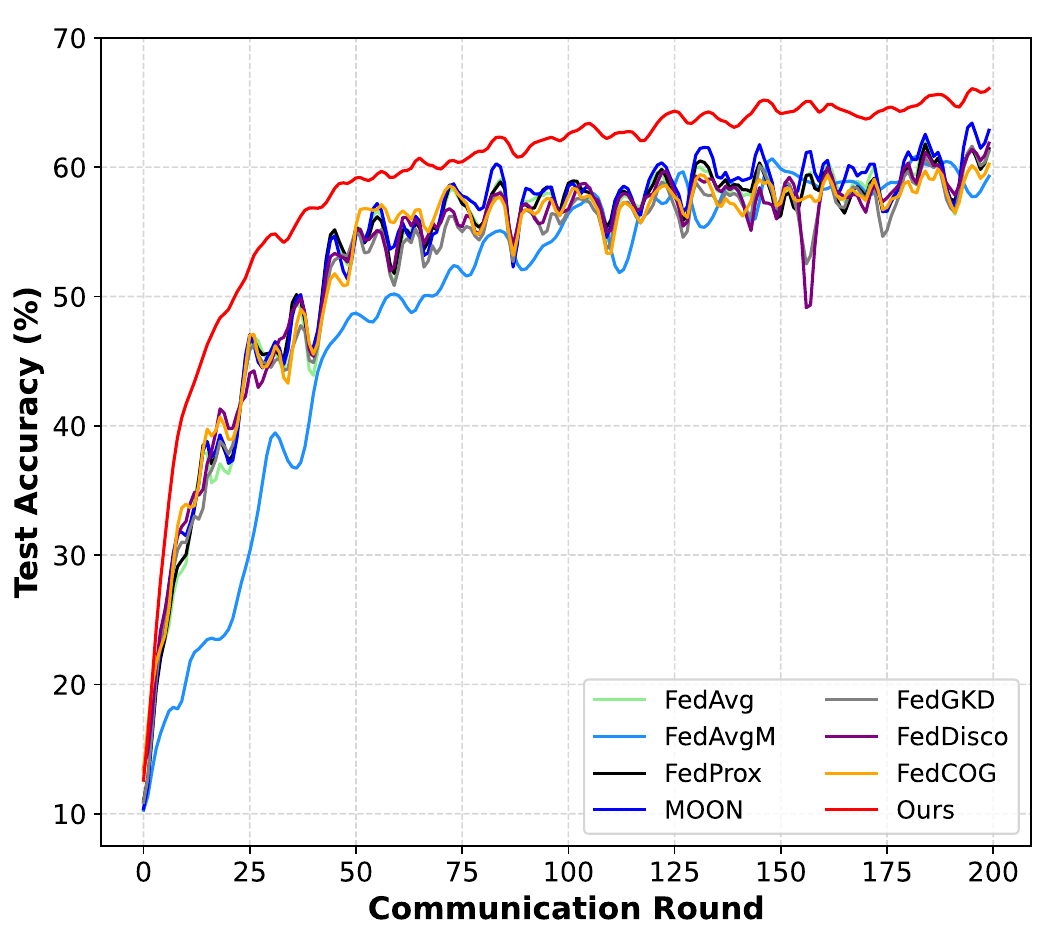}}
		\centerline{\footnotesize (a) CIFAR-10}
	\end{minipage}
	\begin{minipage}{0.32\linewidth}
		\vspace{3pt}
		\centerline{\includegraphics[width=\textwidth]{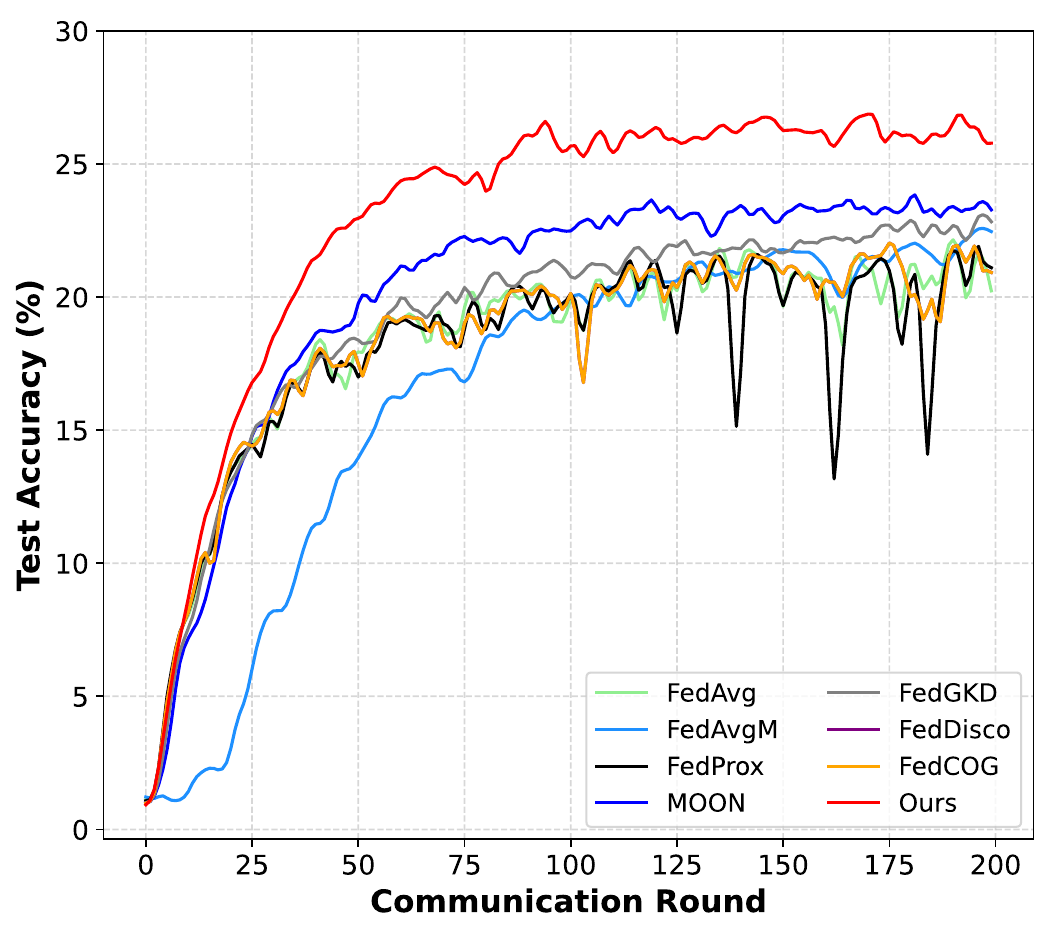}}
		\centerline{\footnotesize (b) CIFAR-100}
	\end{minipage}
        \begin{minipage}{0.32\linewidth}
		\vspace{3pt}
		\centerline{\includegraphics[width=\textwidth]{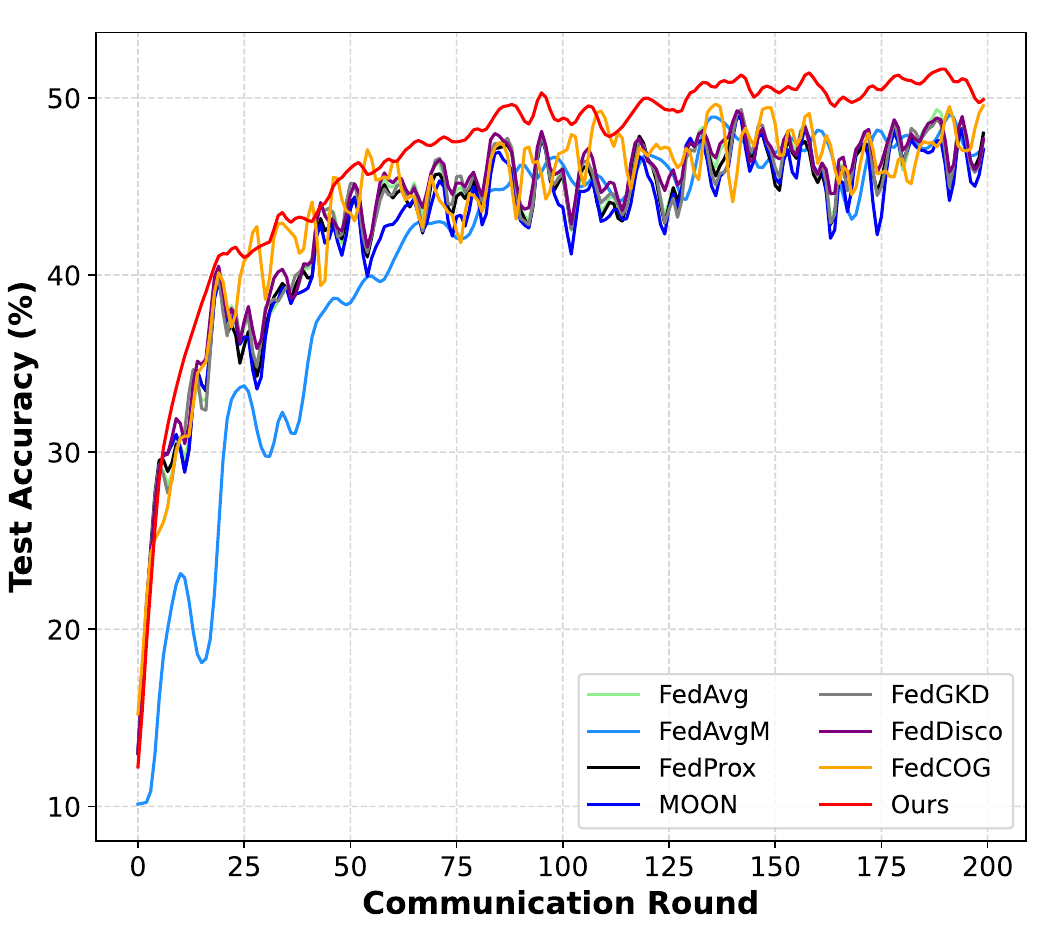}}
		\centerline{\footnotesize (C) CINIC-100}
	\end{minipage}
    \caption{(a)/(b)/(c) show the test accuracy curves for the CIFAR-10, CIFAR-100, and CINIC-10 datasets with $\beta=0.5$, respectively.}
    \label{fig6}
\end{figure*}

\textbf{Teacher weight calculation method}. We conducted experiments on the teacher model aggregation weight calculation method using the CIFAR-10 dataset. The results are shown in Fig. \ref{fig5}. First, we examined the weight distribution of each client using \textit{sinFreqs($F_k^{nums}$)/triFreqs} at different levels of heterogeneity. As shown in Fig. \ref{fig5} (a), since \textit{sinFreqs} is based solely on the data distribution obtained from \textit{Dir($\beta$)}, its distribution becomes more dispersed with higher heterogeneity, and it contains more outliers compared to \textit{tirFreqs}. For \textit{tirFreqs}, there are fewer outliers. At $\beta=0.1$, the distribution of \textit{tirFreqs} is more concentrated compared to \textit{sinFreqs}, but at $\beta=0.5/5.0$, \textit{sinFreqs}' distribution becomes more concentrated as the heterogeneity decreases. As shown in Fig. \ref{fig5} (b), we tracked the variance of three frequencies in \textit{tirFreqs} during training at $\beta=0.5$. Initially, the proportion of client data remains fixed, so its variance does not change. As the frequency of client participation count becomes more evenly distributed during training, the variance gradually decreases to nearly 0. Finally, the variance of the participation interval frequency increases initially and then fluctuates around a certain value. Considering the entire training process (\textit{tirFreqs}), at $\beta=0.1$, the main factors affecting the distribution are data proportion and participation interval frequency. The participation interval frequency helps mitigate the excessive differences in weight distributions caused by data heterogeneity, thus making the aggregation of the teacher model fairer. At $\beta=0.5/5.0$, when the heterogeneity is lower, the participation interval frequency becomes the main factor, so the distribution range is larger than that of \textit{sinFreqs}. This can be beneficial, as the clients participating in training and those not participating need to be distinguished to balance the accumulation of new and old knowledge, avoiding too rapid updates that may cause knowledge forgetting or too slow updates that fail to provide guidance. Overall, \textit{tirFreqs} provides a balance between the knowledge contribution and participation fairness of each client during teacher model aggregation.\\
In Table \ref{tab3}, experiments on \textit{$F_k^{intv}$/$F_k^{part}$/$F_k^{num}$/
triFreqs(AM)/triFreqs(GM)} were conducted under three different data heterogeneity conditions on the CIFAR-10 dataset. \textit{AM} and \textit{GM} represent arithmetic and geometric means, respectively, for calculating $F_k^{tri}$. \textit{GM} performed better under high heterogeneity conditions.

\begin{table}
    \centering
    \caption{ Number of communication rounds ($\#$) required to achieve the highest accuracy of FedAvg and speedup ($\uparrow$). When the required number of rounds is greater than 200, do not perform the calculation on $\uparrow$, use "\textbackslash" instead.}
    \label{tab4}
    \resizebox{0.8\linewidth}{!}{
    \begin{tabular}{c|cccccc}
        \toprule
        \multirow{2}{*}{Method}&\multicolumn{2}{c}{$\beta=0.1$}&\multicolumn{2}{c}{$\beta=0.5$}&\multicolumn{2}{c}{$\beta=5.0$}\\
        \cline{2-7}
        &$\#$&$\uparrow$&$\#$&$\uparrow$&$\#$&$\uparrow$\\
        \cline{1-7}
        &\multicolumn{6}{c}{CIFAR-10 (target acc: 64.07\%)}\\
        \cline{1-7}
        FedAvg \cite{mcmahan2017communication}& 186&1×&184&1×&185&1×\\
        MOON \cite{li2022federated}& $>200$&\textbackslash&137&1.3×&\textbf{154}&\textbf{1.3×}\\
        FedGKD \cite{wang2020tackling}&186&1×&$>200$&\textbackslash&$>200$&\textbackslash\\
        FedDisco \cite{ye2023feddisco}&160&1.2×&$>200$&\textbackslash&$>200$&\textbackslash\\
        Ours&\textbf{73}&\textbf{2.5×}&\textbf{123}&\textbf{1.5×}&161&1.1×\\
        \cline{1-7}
        &\multicolumn{6}{c}{CIFAR-100 (target acc: 22.63\%)}\\
        \cline{1-7}
        FedAvg \cite{mcmahan2017communication}&168&1×&159&1×&194&1×\\
        MOON \cite{li2022federated}&163&$>1\times$&75&2.1×&80&2.4×\\
       FedGKD \cite{wang2020tackling}&168&1×&195&$<1\times$&157&1.2×\\
         FedDisco \cite{ye2023feddisco}&$>200$&\textbackslash&196&$<1\times$&191&$>1\times$\\
        Ours&\textbf{111}&\textbf{1.5×}&\textbf{45}&\textbf{3.5×}&\textbf{43}&\textbf{4.5×}\\
        \cline{1-7}
        &\multicolumn{6}{c}{CINIC-10 (target acc: 50.97\%)}\\
        \cline{1-7}
         FedAvg \cite{mcmahan2017communication}&190&1×&190&1×&176&1×\\
        MOON \cite{li2022federated}&$>200$&\textbackslash&190&1×&\textbf{104}&\textbf{1.7×}\\
        FedGKD \cite{wang2020tackling}&$>200$&\textbackslash&156&1.2×&175&1×\\
         FedDisco \cite{ye2023feddisco}&193&$<1\times$&190&1×&170&1×\\
        Ours&\textbf{111}&\textbf{1.7×}&\textbf{132}&\textbf{1.4×}&121&1.5×\\
        \bottomrule
    \end{tabular}
    }
\end{table}
\begin{table}
    \centering
    \caption{Impact of different number of clients \textit{N} and sampling ratio \textit{C} on experimental results for the CIFAR-10 dataset with $\beta=0.5$.}
    \label{tab5}
    \resizebox{\linewidth}{!}{
        \begin{tabular}{cc|cccccc}
        \toprule
        \multicolumn{2}{c|}{Setting}&\multicolumn{6}{c}{Method}\\
        \cline{1-8}
        \textit{N}&\textit{C}&FedAvg&MOON&FedGKD&FedDisco&FedCOG&Ours(S/T)\\
        \cline{1-8}
        30&0.1&65.12&65.74&64.36&65.66&62.18&66.59/\textbf{69.20}\\
        50&0.1&66.57&66.78&66.77&66.73&63.22&66.17/\textbf{68.10}\\
        100&0.1&64.07&64.84&63.19&62.99&62.09&65.45/\textbf{66.45}\\
        200&0.1&61.21&61.29&61.42&61.78&56.79&61.35/\textbf{62.06}\\
        100&0.05&62.54&62.79&62.73&62.51&62.52&62.62/\textbf{63.89}\\
        100&0.2&65.06&\textbf{66.12}&65.53&65.00&62.46&64.66/64.87\\
        100&0.5&64.61&65.18&64.81&64.90&62.39&\textbf{65.83}/65.70\\
        100&1.0&64.26&64.36&64.22&64.31&62.98&65.24/\textbf{65.29}\\
        \bottomrule
    \end{tabular}
    }
\end{table}
\textbf{Communication round.} We recorded the number of rounds required for FedAvg to reach the highest accuracy and compared the number of rounds ($\#$) and acceleration speed ($\uparrow$) required for MOON, FedGKD, FedDisco, and KDIA to reach the same accuracy. The results are shown in Table \ref{tab4}. KDIA outperforms the baseline methods in most cases, achieving $2.5\times$ acceleration on CIFAR-10 with $\beta=0.1$, and $3\sim4\times$ acceleration on CIFAR-100 with $\beta=0.5/5.0$. This indicates that KDIA exhibits higher convergence speed under more severe heterogeneity. Fig. \ref{fig6} presents the test accuracy curves of KDIA and other baseline methods on the three datasets. KDIA reaches the target accuracy in fewer training rounds, and the convergence process of the teacher model is more stable. In contrast, baseline methods show significant oscillations during training, indicating that under large \textit{N}/small \textit{C} conditions, the training difficulty increases.

\textbf{Number of clients and sampling ratio.} The results are shown in Table \ref{tab5}. For the number of clients, with a fixed sampling ratio $C=0.1$, \textit{N} was selected from $\{30, 50, 100, 200\}$. This setup partially satisfies the "\textit{large N, small C}" condition, and KDIA achieves better performance. The test results align with the analysis in Section \ref{s3}. The baseline methods also follow this trend within a certain range, but an anomaly is observed when $N=30$. This could be due to the fact that, under a fixed \textit{C}, the number of clients \textit{K} involved in training has a greater impact than the gain from clients having more data. For the sampling ratio, with a fixed number of clients $N = 100$, \textit{C} was selected from $\{0.05, 0.1, 0.2, 0.5, 1.0\}$. Under the "\textit{large N}" condition, increasing \textit{C} leads to improvements within a certain range, but the effect diminishes as \textit{C} increases. For example, the result for $C = 0.5$ is worse than for $C = 0.2$, and for KDIA, the results for $C = 0.2/0.5$ are worse than for $C = 0.1$. Therefore, blindly increasing the client sampling ratio is detrimental to the model effect \cite{mcmahan2017communication}, which is consistent with the analysis in Section \ref{s3}.
\begin{table}[h]
    \centering
    \caption{Local time of clients and test accuracy of global model.}
    \label{tab6}
    \begin{tabular}{ccc}
        \toprule
        Method&Local Time (s)&Acc (\%)\\
        \cline{1-3}
        FedAvg \cite{mcmahan2017communication}&$\textbf{56}\pm{9}$&64.07\\
        FedProx \cite{li2020federated}&$59\pm{3}$&64.11\\
        MOON \cite{li2022federated}&$63\pm{3}$&64.84\\
        FedDisco \cite{ye2023feddisco}&$57\pm{11}$&62.99\\
        KDIA&$65\pm{7}$&\textbf{66.45}\\
        \bottomrule
    \end{tabular}
\end{table}
\subsection{Empirical Analysis of KDIA}\label{ss10}
\textbf{Computational cost.} When considering computational cost, federated learning typically places greater emphasis on local clients rather than the server, though the latter's computational cost should not be overlooked. In KDIA, compared to FedAvg, the server incurs additional processes for teacher model aggregation and conditional generator training. For teacher model aggregation, only triple frequency (\textit{triFreqs}) components need updating, followed by geometric averaging and normalization—these steps introduce negligible computational cost. As for conditional generator training, the cost primarily depends on the input image size, the dimensionality of the feature extractor's output space, and the model size, with computational expenses scaling proportionally to these factors. Thus, the server's computational cost mainly stems from conditional generator training. Over 200 global rounds, we measured the average time per round: $118\pm{11}$ seconds.\\
On the client side, KDIA introduces additional steps for generating intermediate features from the conditional generator and knowledge distillation. However, the former incurs minimal overhead—producing features for a batch (size: 64) takes only about 0.011 seconds. We compared the average local training time and final accuracy of FedAvg, FedProx, MOON, FedDisco, and KDIA over 200 rounds, as shown in Table \ref{tab6}. Although KDIA has the highest local training computational cost, it remains within an acceptable range while achieving 1.61\%–3.46\% higher accuracy than other methods \cite{kairouz2021advances}.
\begin{figure}
    \centering
    \includegraphics[width=1.0\linewidth]{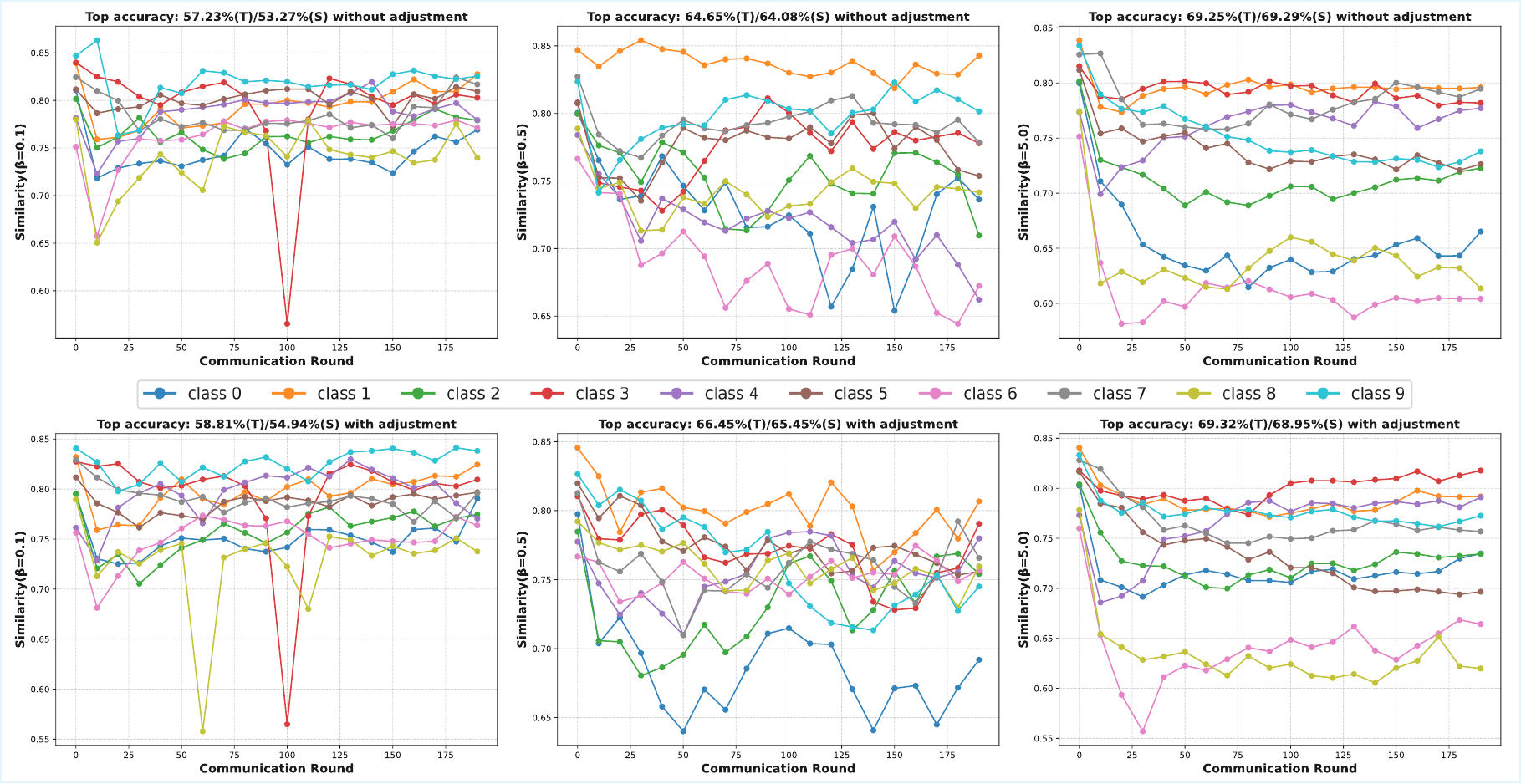}
    \caption{Generated vs. real feature similarity under data heterogeneity in CIFAR-10.}
    \label{fig7}
\end{figure}

\textbf{Communication cost.} In KDIA, during the communication process between the two parties, the server needs to distribute the conditional generator, target model, and teacher model, while clients only upload their local models. Weight computation is performed on the server side, eliminating the need for parameter exchange. Although this introduces two additional rounds of model distribution from the server, it positively enhances local model training performance. When considering global communication cost, as shown in Table \ref{tab4}, KDIA can effectively reduce the number of global rounds required to achieve the target accuracy when only a small subset of clients participate in training, thereby lowering the overall communication cost.

\textbf{Effectiveness of the adjusted conditional generator.} In FedGen \cite{zhu2021data} and FedCG \cite{wu2021fedcg}, Gaussian noise is sampled according to batch size, with sampled labels matching the real data labels. To investigate whether the adjustment method in Section \ref{ss5} is effective, we computed the similarity between real intermediate features and generated features of the same label during training, as well as the final accuracy. The real intermediate features were extracted using a centrally trained model (test accuracy: 81\%). Experiments were conducted under three heterogeneous settings on the CIFAR-10 dataset, with results shown in Fig. \ref{fig7}. The real intermediate features $Z\in{R^{N_{real}\times{C}\times{H}\times{W}}}$ and generated features $\tilde{Z}\in{R^{N_{gen}\times{C}\times{H}\times{W}}}$, where $N_{real}$ and $N_{gen}$ denote the number of features per batch for the same class. For each class, the similarity matrix $S\in{R^{N_{real}\times{N_{gen}}}}$ was computed, and the average similarity was defined as: $\frac{1}{N_{real}\times{N_{gen}}}\sum_{[i,j]}S_{i,j}$.
Experimental results demonstrate that the adjusted generated features align more closely with the real intermediate features. At $\beta=0.1$, both methods perform similarly, but at $\beta=0.5$ and $\beta=5.0$, the adjusted method yields higher overall similarity. For instance, at $\beta=0.5$, 7 classes exceed a similarity of 0.75 with the adjusted method, compared to only 5 classes for the unadjusted version; at $\beta=5.0$, 5 classes exceed 0.75 similarity with adjustment, versus only 4 classes without adjustment. In terms of final accuracy, the adjusted generation method consistently outperforms the unadjusted approach.
\begin{figure}[h]
    \begin{minipage}{0.23\linewidth}
		\vspace{3pt}
		\centerline{\includegraphics[width=\textwidth]{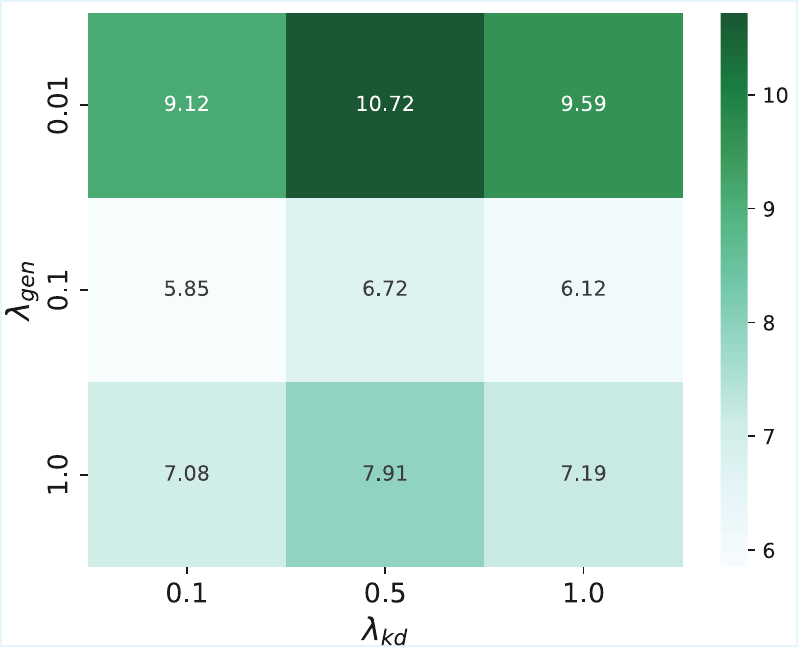}}
		\centerline{\footnotesize (a) $\beta=0.1$}
	\end{minipage}
	\begin{minipage}{0.23\linewidth}
		\vspace{3pt}
		\centerline{\includegraphics[width=\textwidth]{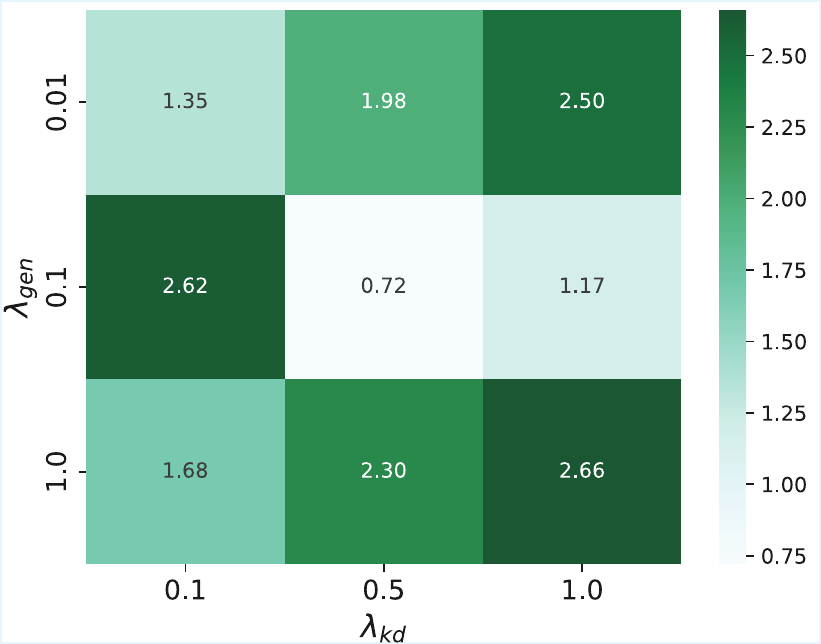}}
		\centerline{\footnotesize (b) $\beta=0.5$}
	\end{minipage}
	\begin{minipage}{0.23\linewidth}
		\vspace{3pt}
		\centerline{\includegraphics[width=\textwidth]{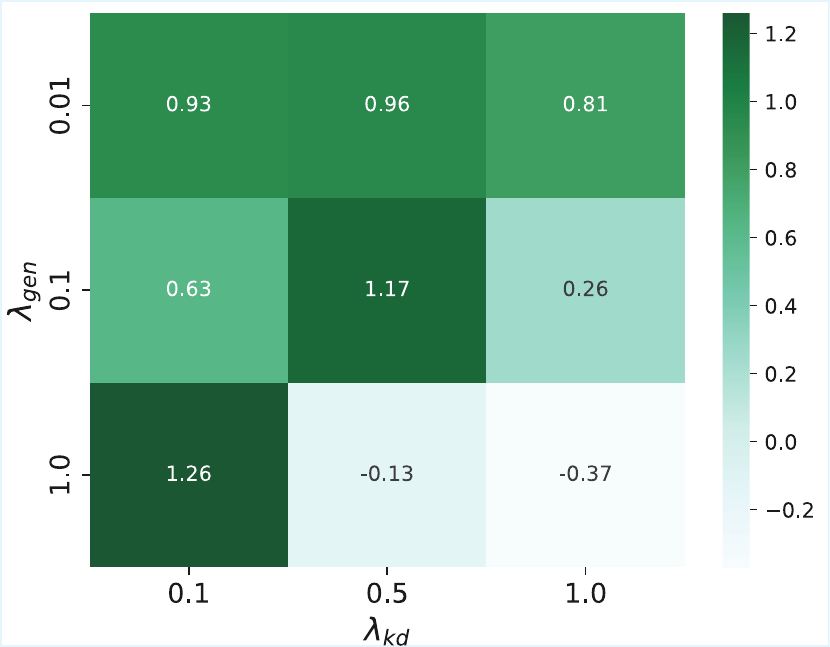}}
		\centerline{\footnotesize (c) $\beta=5.0$}
	\end{minipage}
        \begin{minipage}{0.23\linewidth}
		\vspace{3pt}
		\centerline{\includegraphics[width=\textwidth]{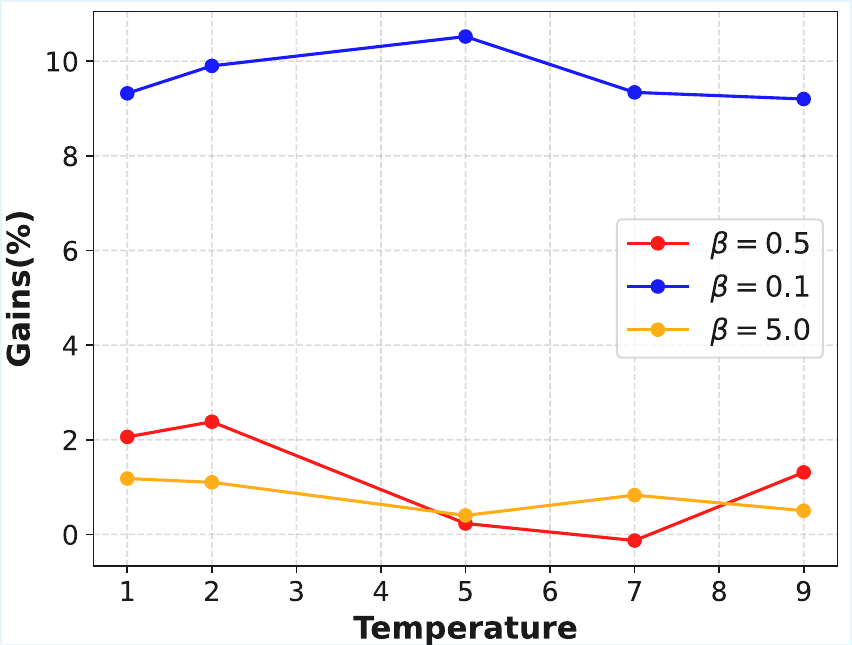}}
		\centerline{\footnotesize (d) $\tau$}
	\end{minipage}
	\caption{On the CIFAR-10 dataset, (a)/(b)/(c) show the accuracy improvements of KDIA relative to FedAvg under different $\lambda_{kd}/\lambda_{gen}$ settings across three levels of data heterogeneity. (d) shows the performance improvements for different distillation temperatures $\tau$ across the three levels of heterogeneity.}
	\label{fig8}
\end{figure}
\subsection{Ablation Study}\label{ss9}
\textbf{Parameter selection ($\lambda_{kd}/\lambda_{gen}/\tau$).} We select $\lambda_{kd}$ and $\lambda_{gen}$ from the sets $\{0.1, 0.5, 1.0\}$ and $\{0.01, 0.1, 1.0\}$, respectively. The test results are shown in Fig. \ref{fig8} (a), (b), and (c). On the CIFAR-10 dataset, KDIA generally outperforms FedAvg in most cases. When $\lambda_{kd}$ is set to 0.5, it performs better in most situations. For simplicity, we fix $\lambda_{kd}$ as 0.5 in all cases. For $\lambda_{gen}$, when $\beta=0.1/0.5/5.0$, the values of $\lambda_{gen}$ are selected to be stable with changes in $\lambda_{kd}$, resulting in $\lambda_{gen}$ values of 0.01/1.0/0.01, respectively. For the distillation temperature $\tau$, we select $\tau$ from the set $\{1.0, 2.0, 5.0, 7.0, 9.0\}$, and the test results are shown in Fig. \ref{fig8} (d). When $\tau$ is set to 2.0, it performs better in all three heterogeneous scenarios. Therefore, $\tau$ is fixed at 2.0 for all experiments.
\begin{table}
    \centering
    \caption{ Ablation study results of KDIA on three datasets with $\beta=0.5$.}
    \label{tab7}
    \begin{tabular}{cccc}
        \toprule
        Method&CIFAR-10&CIFAR-100&CINIC-10\\
        \cline{1-4}
        base (FedAvg)&$64.36\pm{0.69}$&$22.43\pm{0.17}$&$51.19\pm{0.28}$\\
        \cline{1-4}
        \multirow{2}{*}{base+$l_{kd}$}&$64.56\pm{0.26}$&$23.77\pm{0.49}$&	$51.02\pm{0.39}$\\
        &$65.36\pm{0.57}$&$24.88\pm{0.42}$&$51.46\pm{0.48}$\\
        \cline{1-4}
        base+$l_{gen}$&$63.35\pm{0.65}$&$24.67\pm{0.57}$	&$50.89\pm{0.53}$\\
        \cline{1-4}
        \multirow{2}{*}{base+$l_{gen}$+$l_{kd}$}&$65.07\pm{0.39}$	&$25.34\pm{0.54}$&$51.36\pm{0.44}$\\        &\bm{$65.91\pm{0.76}$}&\bm{$26.43\pm{0.60}$}&\bm{$52.22\pm{0.37}$}\\
        \bottomrule
    \end{tabular}
\end{table}

\textbf{self-knowledge distillation (inequitable aggregation) and generated auxiliary data.} We conducted experiments to evaluate the contribution of self-knowledge distillation and generated auxiliary data to KDIA. The results are shown in Table \ref{tab7}. Based on FedAvg (base), we tested the test accuracy of three configurations when $\beta=0.5$: base+$l_{gen}$, base+$l_{kd}$, and base+$l_{kd}$+$l_{gen}$. The base+$l_{kd}$ configuration improved accuracy by 1.0\%, 2.45\%, and 0.27\% on CIFAR-10/100 and CINIC-10, respectively, showing a significant contribution. Although base+$l_{gen}$ only showed improvement on CIFAR-100, combining $l_{kd}$ and $l_{gen}$ (base+$l_{kd}$+$l_{gen}$) resulted in even greater performance gains. In self-knowledge distillation, \textit{triFreqs} helps the teacher model better guide the student model during training, and when the local model performance improves, it also assists in guiding the server-side generator’s training process, which in turn generates better auxiliary data for training. This explains why base+$l_{gen}$ did not perform well, but base+$l_{kd}$+$l_{gen}$ achieved a greater performance boost.
\section{Conclusion and Future works}
Data heterogeneity is a significant issue in federated learning, as the differences in data distributions across clients cause the aggregated global model to deviate from the target direction. Furthermore, a large number of clients and low sampling ratios increase the difficulty of the learning task. To address this scenario, which is characteristic of federated learning, we propose the KDIA method. First, we use \textit{triFreqs} to aggregate all clients and obtain a teacher model. Then, the teacher model guides a subset of student models for local training. This approach leverages the knowledge of all clients while only involving a small fraction of them, thus improving model performance and reducing communication costs. Additionally, a conditional generator is used to generate locally distributed data that approximates a uniform class distribution to assist in training. Extensive experiments demonstrate the effectiveness of KDIA. In future work, we will validate the algorithm's effectiveness on more realistic datasets and under broader heterogeneity scenarios.






\end{document}